\documentclass[letterpaper]{article} 
\usepackage{aaai2026}  
\usepackage{times}  
\usepackage{helvet}  
\usepackage{courier}  
\usepackage[hyphens]{url}  
\usepackage{graphicx} 
\usepackage{natbib}  
\usepackage{caption} 
\usepackage{newfloat}
\usepackage{listings}

\usepackage{amsmath}
\usepackage{amssymb}
\usepackage{xcolor}
\usepackage{booktabs}
\usepackage{multirow}
\usepackage{subcaption}
\usepackage{colortbl}
\usepackage{array} 
\usepackage{soul}    %
\usepackage{enumitem}
\definecolor{darkgreen}{RGB}{0, 100, 0}
\usepackage[ruled,vlined]{algorithm2e}
\usepackage{tabularx}
\usepackage{makecell}
\usepackage{cleveref} 
\newcolumntype{P}[1]{>{\centering\arraybackslash}p{#1}}
\usepackage[most]{tcolorbox}

\tcbset{   
  promptbox/.style={
    colback=gray!10,
    colframe=gray!80,
    sharp corners,
    boxrule=0.5pt,
    left=4pt,
    right=4pt,
    top=2pt,
    bottom=2pt,
    fonttitle=\bfseries,
    title=System Prompt
  }
}

\definecolor{lightred}{RGB}{255,150,150}
\definecolor{transparentred1}{RGB}{255,220,220}
\definecolor{transparentred2}{RGB}{255,180,180}
\definecolor{transparentred3}{RGB}{255,140,140}

\definecolor{lightgreen}{RGB}{150,200,150}
\definecolor{transparentgreen1}{RGB}{214,245,214}
\definecolor{transparentgreen2}{RGB}{170,230,170}
\definecolor{transparentgreen3}{RGB}{130,190,130}

\definecolor{transparentorange1}{RGB}{255,228,181}
\definecolor{transparentorange2}{RGB}{255,210,143}
\definecolor{transparentorange3}{RGB}{255,185,0}

\newcommand{\hlred}[2]{%
  \ifnum#2<5
    \sethlcolor{transparentred1}%
  \else
    \ifnum#2<10
      \sethlcolor{transparentred2}%
    \else
      \sethlcolor{transparentred3}%
    \fi
  \fi
  \hl{#1}%
}

\newcommand{\hlgreen}[2]{%
  \ifnum#2<5
    \sethlcolor{transparentgreen1}%
  \else
    \ifnum#2<10
      \sethlcolor{transparentgreen2}%
    \else
      \sethlcolor{transparentgreen3}%
    \fi
  \fi
  \hl{#1}%
}

\usepackage{calc}
\newcommand{\hlorange}[2]{%
  \ifdim#2pt<10pt 
    \sethlcolor{transparentorange1}%
  \else
    \ifdim#2pt<50pt
      \sethlcolor{transparentorange2}%
    \else
      \sethlcolor{transparentorange3}%
    \fi
  \fi
  \hl{#1}%
}

\DeclareCaptionStyle{ruled}{labelfont=normalfont,labelsep=colon,strut=off} 
\nocopyright 
\title{\textsc{SCoPe}: Intrinsic Semantic Space Control for Mitigating Copyright Infringement in LLMs}
\author{
    Zhenliang Zhang\textsuperscript{\rm $1$,$2$},\
    Xinyu Hu\textsuperscript{\rm $1$},\
    Xiaojun Wan\textsuperscript{\rm $1$,}\thanks{Corresponding author.}
}
\affiliations{
    \textsuperscript{\rm 1}Wangxuan Institute of Computer Technology, Peking University\\
    \textsuperscript{\rm 2}School of Software and Microelectronics, Peking University

    \texttt{zhenliang@stu.pku.edu.cn},\
    \texttt{\{huxinyu,wanxiaojun\}@pku.edu.cn}
}

\usepackage{bibentry}

\begin{document}

\maketitle

\begin{abstract}
Large language models sometimes inadvertently reproduce passages that are copyrighted, exposing downstream applications to legal risk.
Most existing studies for inference-time defences focus on surface-level token matching and rely on external blocklists or filters, which add deployment complexity and may overlook semantically paraphrased leakage.  
In this work, we reframe copyright infringement mitigation as \textbf{intrinsic semantic-space control} and introduce \textsc{SCoPe}, an inference-time method that requires no parameter updates or auxiliary filters.  
Specifically, the sparse autoencoder (SAE) projects hidden states into a high-dimensional, near-monosemantic space; benefiting from this representation, we identify a copyright-sensitive subspace and clamp its activations during decoding.  
Experiments on widely recognized benchmarks show that \textsc{SCoPe} mitigates copyright infringement without degrading general utility. Further interpretability analyses confirm that the isolated subspace captures high-level semantics.
\end{abstract}

\section{Introduction}
\label{sec:intro}
Large language models (LLMs) have demonstrated impressive capabilities in generating high-quality content. However, a surge of copyright lawsuits from media organizations and creators has raised serious legal concerns, particularly regarding the potential reproduction of copyrighted material from training data \cite{yu2023codeipprompt, duan2024ai, brunettitraining, stratton2024market}. Consequently, mitigating copyright infringement in LLMs has emerged as a pressing and essential research challenge.

Existing defenses against copyright infringement generally  fall into three research directions: preprocessing filters to exclude copyrighted material \cite{ kandpal2022deduplicating, sag2023copyright}, training-time interventions such as Near Access-Freeness and selective unlearning \cite{abad2024copyright, xu2025suv}, and inference-time controls. While preprocessing and training-time methods offer structural guarantees, they are computationally intensive and lack post-deployment flexibility---particularly when new protected content emerges. In contrast, inference-time techniques offer adaptability without modifying model parameters, making them more attractive and practical in real-world settings.


However, most prior inference-time methods rely heavily on external artifacts such as blocklist corpora or bloom filters \cite{ippolito2022preventing, shi-etal-2024-trusting}, which require costly string-level comparisons or real-time rewriting. These mechanisms increase system complexity and may degrade fluency \cite{zhang2025certified}. This motivates a question: \textit{Can we eliminate these external mechanisms and enable LLMs to intrinsically avoid generating infringing content?}

Therefore, \textbf{we shift from surface-level output filtering to intrinsic semantic space control}. 
Inspired by the semantic subspace hypothesis \cite{ park2023representation, ferrando2024know}, we consider the possibility that copyrighted content may correspond to a distinct and identifiable subspace within the representation space of LLMs. 
If such a copyrighted subspace can be identified and effectively suppressed during LLM inference, infringement can be mitigated intrinsically.
\begin{quotation}

\noindent
\textit{“Once the mind ordains its inner limits, outward conduct is thereby governed.”}

\raggedleft \textit{—~Record of Rites}
\end{quotation}

The primary technical challenge arises from the \textbf{polysemanticity} of LLM neurons: individual neurons often encode multiple concepts~\cite{olah2020zoom}. 
This semantic entanglement makes it difficult to isolate subspaces specifically associated with copyrighted content. Consequently, two critical research questions emerge: identifying copyright-sensitive subspaces and intervening in them to achieve effective copyright protection.

To address these challenges, we propose \textsc{SCoPe}, a two-stage method built on the sparse autoencoder (SAE). In the first stage, dense hidden states are mapped into a high-dimensional sparse space with \textbf{monosemanticity}. Leveraging this property, we define the ideal copyrighted subspace and develop a practical algorithm to estimate it.
In the second stage, we apply feature clamping during decoding to suppress activations in this subspace, thereby reducing the risk of reproducing protected content. Experimental results show that, compared to baselines, \textsc{SCoPe} achieves superior copyright protection performance while maintaining general utility.

Moreover, we provide empirical validation of the subspace hypothesis by demonstrating that sparse space enable clearer separation between copyrighted and general content. Through feature semantic interpretation and reverse interventions, we confirm that the copyrighted subspace achieves isolation at the semantic level. 

Our contributions are summarized as follows:


\begin{itemize}
  \item \textbf{Novel Subspace Perspective:} We frame copyrighted content as residing in a distinct semantic subspace of LLM representations and develop an efficient method to identify the copyrighted subspace.
  \item \textbf{Semantic-Level Mitigation:} Our \textsc{SCoPe} framework clamps high-risk semantic features at decode time without external filters, operating at the semantic level rather than via surface token matching. 
  \item \textbf{Effectiveness and Interpretability:} \textsc{SCoPe} delivers substantial reductions in copyright leakage while preserving overall model utility, with semantically meaningful controls. 
\end{itemize}

\section{Preliminaries}

\subsection{Semantic Space in LLMs}
In LLMs, each layer produces a hidden state (embedding vector) $\mathbf{h}\in\mathbb{R}^d$ for every input token.  The collection of all possible hidden states at a given layer therefore defines a $d$‐dimensional \textbf{semantic space}, which serves as the model’s internal representation of meaning at that processing stage.

\paragraph{Polysemanticity}
It is a well-documented phenomenon in these spaces: individual neurons often encode multiple, unrelated concepts simultaneously. Such superposition of neurons makes direct interpretation of single dimensions difficult \cite{olah2020zoom, elhage2022superposition}.  As a result, methods that seek to identify and control specific semantic attributes must first disentangle these overlapping representations into more sparse subspaces.

\subsection{SAE and Sparse Space} 
\label{sec:sae}

Sparse Autoencoder (SAE) is a neural module that projects the dense hidden states of an LLM into a higher-dimensional, sparse semantic space. Its core motivation is to address \emph{polysemanticity} by mapping activations into disentangled, interpretable directions.

Given a hidden state vector $\mathbf{h}\in\mathbb{R}^d$, the SAE applies an \textbf{encoder} $f:\mathbb{R}^d\to\mathbb{R}^k$ and a \textbf{decoder} $g:\mathbb{R}^k\to\mathbb{R}^d$, with $k\gg d$:
\begin{align}
\mathbf{z} &= f(\mathbf{h}) = \mathrm{JumpReLU}(\mathbf{W}_{\mathrm{enc}}\mathbf{h} + \mathbf{b}_e)\, \label{eq:sae_encoder}\\
\hat{\mathbf{h}} &= g(\mathbf{z}) = \mathbf{W}_{\mathrm{dec}}\mathbf{z} + \mathbf{b}_d\, \label{eq:sae_decoder}
\end{align}
Here, $\mathbf{z}\in\mathbb{R}^k$ is the sparse, high‐dimensional vector, and $\hat{\mathbf{h}}$ is the reconstructed hidden state fed back into the LLM’s residual stream.  $\mathrm{JumpReLU}(x)$ is a variant of ReLU that enforces strict sparsity by zeroing out small activations, with the detailed formulations included in Appendix~A.
Moreover, the training of SAEs is not complicated, and some have been publicly available, for instance, the GemmaScope family \cite{lieberum2024gemmascopeopensparse}.
\paragraph{Monosemanticity}
The vector \(\mathbf{z}\) defines a \(k\)-dimensional sparse semantic space.  To enforce sparsity, the SAE is trained with a reconstruction objective augmented by a sparsity penalty, yielding a disentangled, interpretable encoding of the original hidden state.  Empirical studies confirm that, activations from SAE exhibit \textbf{monosemanticity} rather than semantic superposition \cite{cunningham2023sparse,pach2025sparse}.  
This architecture enables each dimension of \(\mathbf{z}\) to correspond to distinct narrative or thematic elements (e.g., dialogue, temporal markers, topic shifts), enabling interpretation and targeted intervention.



\subsection{SAE-Induced Semantic Subspaces} 
\label{sec:subspace_in_sae}

Given a hidden state vector \(\mathbf{h}\in\mathbb{R}^d\) from an LLM, we obtain its sparse encoding \(\mathbf{z}=[z_1,\dots,z_k]\in\mathbb{R}^k\) via a pretrained SAE. The resulting sparse semantic space is designed so that each dimension \(z_i\) captures disentangled concepts. 

More formally, for any index set of dimensions \(\mathcal{I}\subseteq\{1,\dots,k\}\), we define the projection:
\begin{equation}
  \bigl(P_{\mathcal{I}}(\mathbf{z})\bigr)_i =
  \begin{cases}
    z_i, & i \in \mathcal{I}\\
    0,   & i \notin \mathcal{I}
  \end{cases}
\end{equation}
Then the \textbf{semantic subspace} associated with \(\mathcal{I}\) is 
\begin{equation}
\mathcal{S} = \mathcal{S}(\mathcal{I})
=\bigl\{\mathbf{z}\in\mathbb{R}^k \mid P_{\mathcal{I}}(\mathbf{z})=\mathbf{z}\bigr\}
\end{equation}
namely the set of sparse activations supported only on the chosen dimensions.

\paragraph{Subspace Hypothesis}
Evidence suggests that distinct semantic attributes (such as gender bias, unknown entity, or syntax) tend to align with specific directions in embedding spaces \cite{bolukbasi2016man, park2023representation, ferrando2024know}.  By extension, we hypothesize that the SAE-induced sparse space assigns most interpretable concepts to distinct linear subspaces. Consequently, activations for each concept concentrate within a subspace $\mathcal{S}$, allowing us to isolate and control it simply by projecting onto or away from $\mathcal{S}$.
\section{Methodology}
\label{sec:methodology}


Building on the subspace hypothesis, we introduce \textbf{S}ubspace-oriented \textbf{Co}pyright \textbf{P}rot\textbf{e}ction (\textbf{\textsc{SCoPe}}), a two-stage method for LLM infringement prevention.  
In the first stage (Section~\ref{sec:copyright_subspace_1}), we identify the \textbf{ copyrighted subspace}, whose constituent dimensions activate preferentially on copyrighted content. 
In the second stage (Section \ref{sec:scope}), we clamp this subspace at inference time to suppress copyright-sensitive outputs.

\subsection{Identifying the Copyrighted Subspace}
\label{sec:copyright_subspace_1}

We center our investigation on three core questions:
(1)~\textit{\textbf{Target}}: What is the ideal copyrighted subspace? (2)~\textit{\textbf{Distance}}: How do we quantify a candidate subspace’s deviation from that ideal? (3)~\textit{\textbf{Approach}}: How do we empirically refine our subspace to better approximate the target? 

\begin{figure*}[th]
  \centering
  \begin{subfigure}[h]{0.65\textwidth}
    \centering
    \includegraphics[width=\linewidth]{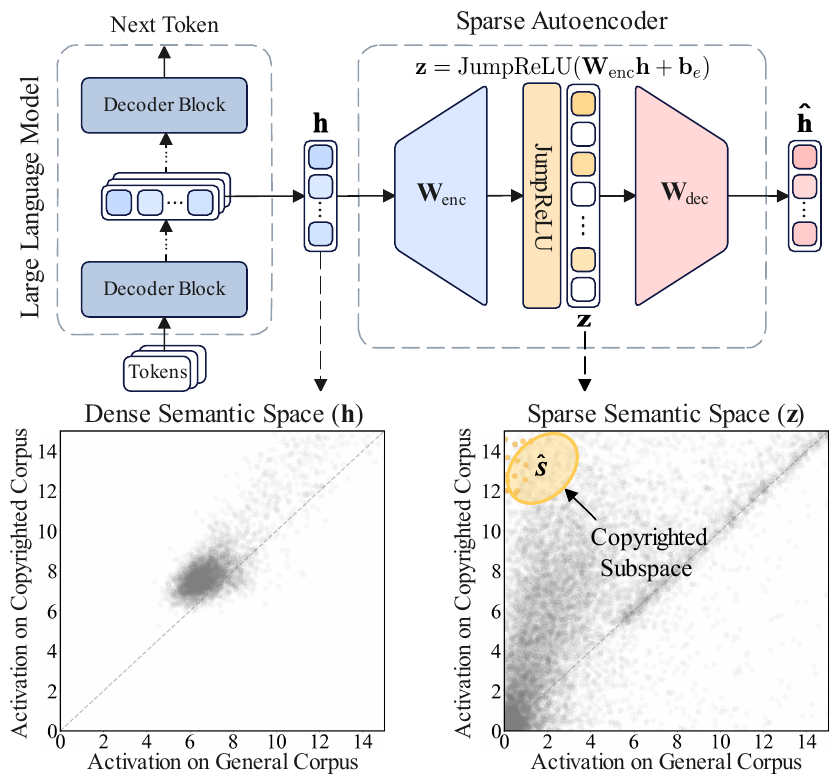}
    \caption{\textbf{Activation in dense vs.\ sparse semantic spaces.}
      Points represent activations of dimensions in the LLM dense space (left) and SAE-induced sparse space (right).}
    \label{fig:dense_sparse_space}
  \end{subfigure}
  \hfill
  \begin{minipage}[h]{0.34\textwidth}
    \centering
    \begin{subfigure}[h]{\textwidth}
      \includegraphics[width=\linewidth]{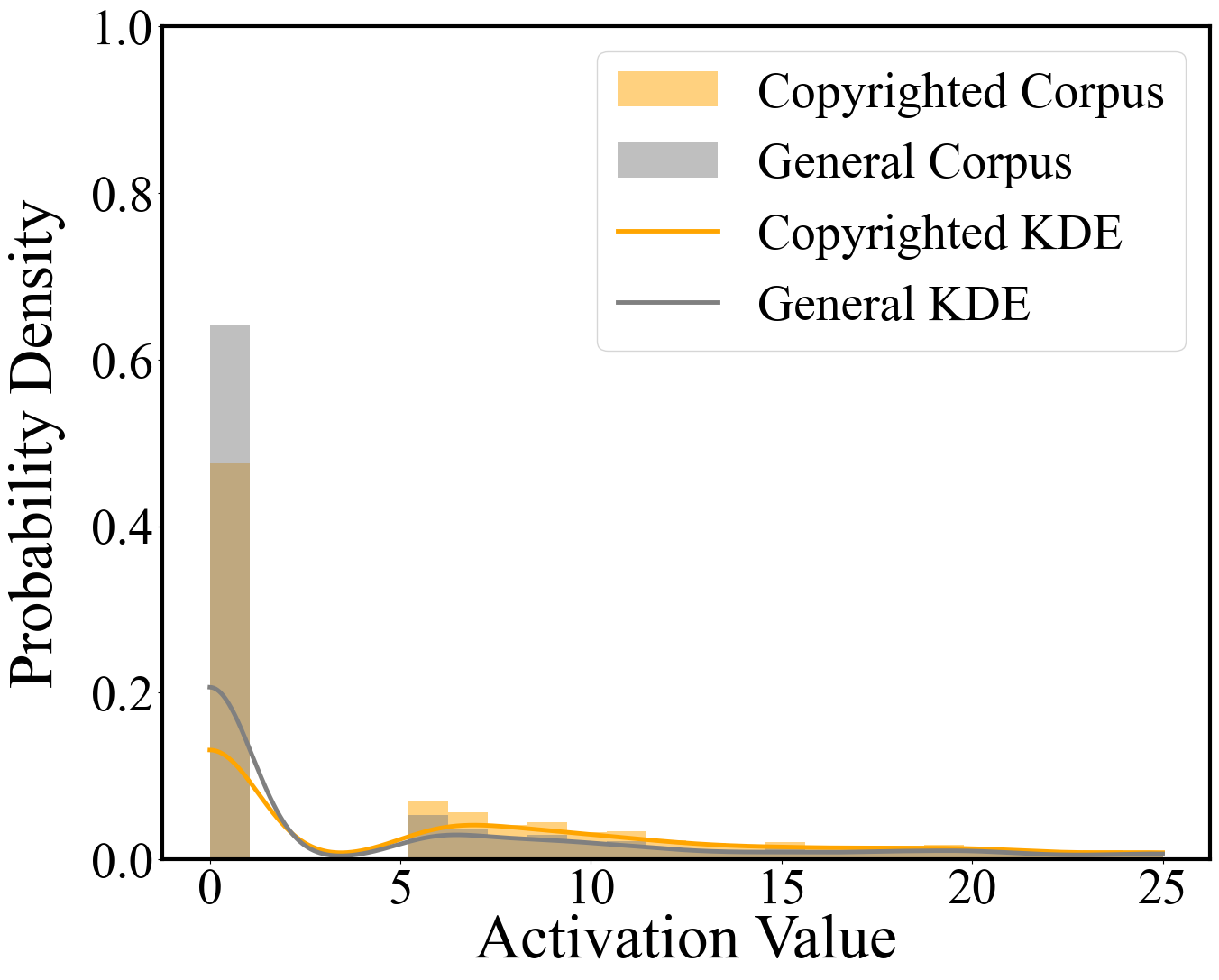}
      \caption{Activation in full semantic space.}
      \label{fig:whole_space_distribution}
    \end{subfigure}
    \vspace{-1.2ex}
    \begin{subfigure}[h]{\textwidth}
      \includegraphics[width=\linewidth]{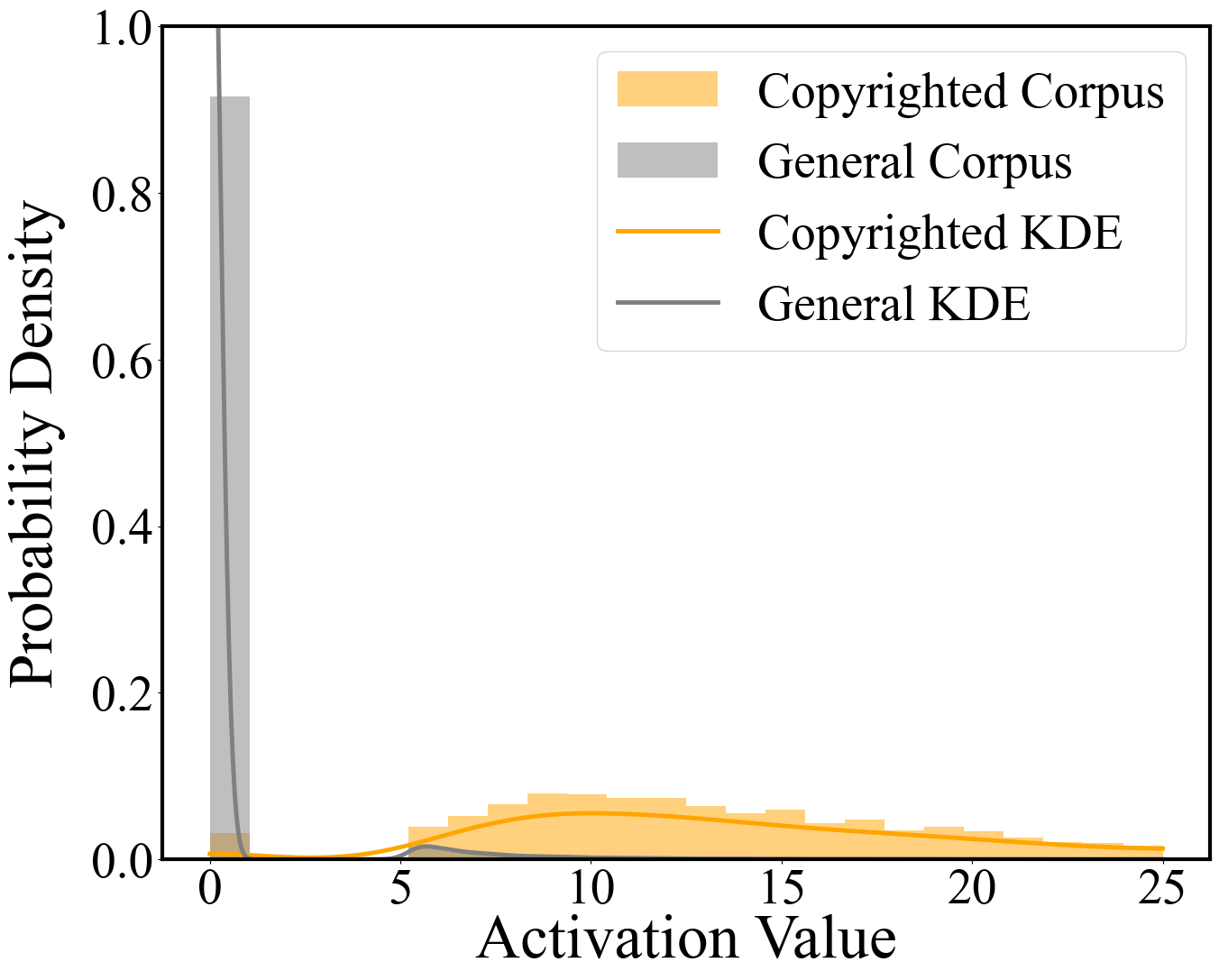}
      \caption{Activation in estimated subspace \( \hat{\mathcal{S}} \).}
      \label{fig:copyright_subspace_distribution}
    \end{subfigure}
  \end{minipage}

  \caption{\textbf{Visualization of semantic separation and subspace discrimination.}
    (a) SAE sparse space enables better separation of copyright-sensitive dimensions. 
    (b) In the full space, activations for both corpora overlap. 
    (c) In the estimated subspace \( \hat{\mathcal{S}} \), activations become clearly separable.
  }
  \label{fig:em_study}
\end{figure*}

\subsubsection{Ideal Copyrighted Subspace (Target)}
\label{sec:ideal-empirical}

We aim to identify a subspace that captures \textit{only and exactly} the semantics of copyrighted corpus. Formally, let \( \mathcal{C}_{\mathrm{cr}} \) and \( \mathcal{C}_{\mathrm{gen}} \) denote the copyrighted and general corpora, respectively.
Each input text $x\in\mathcal{C}$ is encoded by the SAE into a sequence of $k$-dimensional vectors. We then apply token-level max pooling over that sequence to obtain the final sparse vector $\mathbf{z}^{(x)}\in\mathbb{R}^k$. The ideal copyrighted subspace $\mathcal{S}^*$ should correspond to 
an index set \( \mathcal{I}^* \) that satisfies the following two properties:

\begin{enumerate}
\item Coverage: For each sample \(x_{\mathrm{cr}}\in\mathcal{C}_{\mathrm{cr}}\), each dimension \(i\in\mathcal{I}^*\) of $\mathbf{z}^{(x_{\mathrm{cr}})}$ should exceed an activation threshold:
\begin{equation}
   z^{(x_{\mathrm{cr}})}_i > \tau, \quad
  \forall\,x_{\mathrm{cr}} \in \mathcal{C}_{\mathrm{cr}}, \ i \in \mathcal{I}^*
  \label{eq:ideal-cov-revised}
\end{equation}
\item Exclusivity: For each \(x_{\mathrm{gen}}\in\mathcal{C}_{\mathrm{gen}}\), each dimension \(i\in\mathcal{I}^*\) of $\mathbf{z}^{(x_{\mathrm{gen}})}$ should be zero:
\begin{equation}
 z^{(x_{\mathrm{gen}})}_i = 0, \quad
\forall\,x_{\mathrm{gen}} \in \mathcal{C}_{\mathrm{gen}},\ i \in \mathcal{I}^*
\label{eq:ideal-exc}
\end{equation} 
\end{enumerate}

\subsubsection{Copyright Alignment Score (Distance)}
\label{sec:subspace-estimation}

Although the optimal subspace $\mathcal{S}^*=\mathcal{S}(\mathcal{I^*})$ defines an ideal target, it is unattainable in practice due to potential residual polysemanticity, even after the SAE transformation.

Consequently, we loosen the strict Coverage and Exclusivity constraints and introduce an empirical metric, called \textbf{Copyright Alignment Score}, to assess the quality of any candidate subspace $\mathcal{S}$:
\begin{align}
\small
    \mathcal{Q}(\mathcal{S}) 
= 
\mathbb{E}_{\substack{i \sim \mathcal{I} \\ x_{\mathrm{cr}} \sim \mathcal{C}_{\mathrm{cr}} \\ x_{\mathrm{gen}} \sim \mathcal{C}_{\mathrm{gen}}}}
\left[ 
  \mathbb{I}\left( z_i^{(x_{\mathrm{cr}})} > z_i^{(x_{\mathrm{gen}})} \right) 
\right]
\label{eq:subspace_score}
\end{align}

This score measures the probability that a randomly selected dimension in \(\mathcal{S}\) has a higher activation on a copyrighted sample than on a general sample. In other words, it captures the degree to which \(\mathcal{S}\) preferentially responds to \(\mathcal{C}_{\mathrm{cr}}\) over \(\mathcal{C}_{\mathrm{gen}}\). 

It behaves as follows in two extreme cases:
\begin{itemize}
  \item \textbf{Ideal subspace:} If \(\mathcal{S} = \mathcal{S}^*\), then each dimension in \(\mathcal{S}\) activates exclusively for copyrighted inputs.  Hence for all \(i \in \mathcal{I}^*\), $ z_i^{(x_{\mathrm{cr}})} > z_i^{(x_{\mathrm{gen}})}$,
  yielding \(\mathcal{Q}(\mathcal{S}^*) = 1\).
  \item \textbf{Neutral subspace:} If \(\mathcal{S}\) lacks preferential information, so that activations for \(\mathcal{C}_{\mathrm{cr}}\) and \(\mathcal{C}_{\mathrm{gen}}\) are drawn from the same distribution, then \(\mathcal{Q}(\mathcal{S}) \approx 0.5\), indicating that there is no preference in activation.
\end{itemize}

\subsubsection{Constructing the Empirical Copyrighted Subspace (Approach)}
\label{sec:construct_subspace}

Exhaustively evaluating \(\mathcal{Q}(\mathcal{S})\) over all \(2^{k}\) possible subspaces is practically infeasible.  
We therefore simplify this computation by using the following upper bound (proof included in Appendix~B.2):
\begin{align}
    \mathcal{Q}(\mathcal{S}) \leq \max_{i \in \mathcal{I}} \mathcal{Q}(i)
\label{eq:score-bound}
\end{align}

Here, \(\mathcal{Q}(i)\) is a shorthand for \(\mathcal{Q}\bigl(\mathcal{S}(\{i\})\bigr)\).
This bound implies that the score of any subspace is upper-bounded by the best individual dimension. 
Consequently, adding any lower-scoring dimension cannot increase (and often reduces) the overall alignment score.  
This observation turns subspace search into the simpler problem of identifying high-scoring single dimensions.

We rank all dimensions by their Copyright Alignment Scores \(\mathcal{Q}(i)\) and keep the top \(n\):
\begin{align}
    \hat{\mathcal{I}}
=\bigl\{\,i \mid \mathcal{Q}(i)\ge\theta_{n}\bigr\},
\end{align} 
where \(\theta_n\) is the cutoff equal to the \(n\)-th largest score.
While this greedy selection may not be optimal when activations are correlated, it still offers a reliable approximation and runs in linear time. The parameter \( n \) controls a trade-off between subspace compactness and approximation fidelity. And the \textbf{empirical copyrighted subspace} is $\hat{\mathcal{S}} = \mathcal{S}(\hat{\mathcal{I}})$.

\subsection{Validation of the Copyrighted Subspace}

Before detailing the protection mechanism, we empirically validate the quality and behavior of the estimated subspace \( \hat{\mathcal{S}} \).
We begin by validating two key questions of our approach:  
(1) \textbf{Dense Space \(\rightarrow\) Sparse Space}, whether the SAE-transformed space contains localized dimensions responsive to copyrighted material;  
(2) \textbf{Sparse Space \(\rightarrow\) Copyrighted Subspace}, whether the estimation method in Section~\ref{sec:subspace-estimation} can effectively identify such a subspace from data. Experimental setup is detailed in Section~\ref{sec:exp_setup} and Appendix~C.1.

\paragraph{Semantic Separation in Sparse Space}
We evaluate whether SAE facilitates the separation of copyrighted information by comparing dimension-wise activations across corpora. For each semantic dimension, we compute its average activation on the general (x-axis) and copyrighted (y-axis) corpora, and plot the results in Figure~\ref{fig:dense_sparse_space}. Each point represents one dimension in the semantic space.

In the original LLM space (Figure~\ref{fig:dense_sparse_space} left), most dimensions cluster near the diagonal \( y = x \), reflecting similar average activations across the two corpora. In contrast, the SAE-Induced space (Figure~\ref{fig:dense_sparse_space} right) exhibits greater dispersion, with some dimensions appearing in the upper-left quadrant--indicating strong activation on copyrighted content and minimal activation on general content.

\paragraph{Effectiveness of Empirical Subspace Estimation}
We evaluate the quality of the estimated subspace \( \hat{\mathcal{S}} \) by comparing activation distributions in the full sparse semantic space and in \( \hat{\mathcal{S}} \) itself.

As shown in Figure~\ref{fig:whole_space_distribution}, activations across general and copyrighted corpora are highly overlapping in the full space, indicating poor separability. In contrast, Figure~\ref{fig:copyright_subspace_distribution} shows that within \( \hat{\mathcal{S}} \), copyrighted inputs consistently activate the selected dimensions, while general content remains largely inactive. Both plots visualize the activation threshold $\tau=5$ (Details are provided in Appendix~C.2).

\paragraph{\textbf{Takeaways}}  
SAE disentangles semantic dimensions, making copyright-sensitive dimensions more localized and separable. The estimated subspace \(\hat{\mathcal{S}}\) exhibits strong semantic selectivity, empirically validating our Subspace Hypothesis (Section~\ref{sec:subspace_in_sae}) and demonstrating the effectiveness of isolating copyright-sensitive dimensions.

\subsection{Copyrighted Subspace Protection}
\label{sec:scope}

In the second stage, 
\textsc{SCoPe} operates directly on the sparse semantic representation to prevent the model from reproducing protected content.

We implement a simple yet effective mechanism called \textbf{feature clamping} \cite{bricken2023monosemanticity}. At each decoding step, we first project the hidden state \( \mathbf{h} \in \mathbb{R}^d \) into the sparse semantic space via the pretrained SAE encoder:  
$\mathbf{z} = f(\mathbf{h}) \in \mathbb{R}^k$.
We apply feature clamping by modifying each semantic dimension \( z_i \) in the sparse activation vector \( \mathbf{z} \). For any dimension in the copyrighted subspace \( \hat{\mathcal{S}} \), if its activation exceeds a threshold \( \tau \), we suppress it to zero:
\begin{equation}
  z_i \leftarrow
  \begin{cases}
    0, & \text{if } i \in \hat{\mathcal{I}} \text{ and } z_i > \tau\\
    z_i, & \text{otherwise}
  \end{cases}
\label{eq:feature-clamping}
\end{equation}
The operation ensures that only activated dimensions within the subspace are suppressed, preserving general semantics outside \( \hat{\mathcal{S}} \).

After clamping, we reconstruct $\hat{\mathbf{h}}$ via SAE decoder $ \hat{\mathbf{h}}=g(\mathbf{z}) \in \mathbb{R}^d$ and add the reconstruction error, then fed back into the LLM’s residual stream, ensuring that all subsequent decoder generate from the suppressed subspace.

\paragraph{Interpretation}
Unlike previous approaches that rely on surface-level similarity, such as $n$-gram or vector embedding comparisons \cite{ippolito2022preventing,wei2024evaluating}, our method achieves \textbf{semantic-level isolation} by directly suppressing neural activations within a subspace linked to copyrighted content. This mechanism steers the model's generation away from risky semantic subspace, while leaving unrelated conceptual intact.
And it is lightweight, integrates seamlessly into decoding, and offers interpretability.


\begin{table*}[!ht]
\centering
\small
\setlength{\tabcolsep}{2pt}  
\begin{tabular}{@{} 
    P{2cm}   
    P{3cm}   
    P{1.7cm}   
    P{1.7cm}   
    P{1.7cm}   
    P{1.7cm}   
    P{1.6cm}   
    P{1.6cm}   
@{}}
\toprule
 &  & \multicolumn{4}{c}{\textbf{Win Rate on Mitigation Metrics ($\uparrow$, \%)}} 
       & \multicolumn{2}{c}{\textbf{Utility Preservation ($\uparrow$)}} \\
\cmidrule(l){3-8}
\multirow{-2}{*}{\textbf{Model}} 
  & \multirow{-2}{*}{\textbf{Method}} 
  & Semantic Similarity 
  & MinHash Similarity 
  & Levenshtein Distance 
  & \cellcolor[HTML]{EFEFEF}\textbf{ Average win rate} 
  & Blocklisted F1 
  & In-Domain F1 \\
\midrule
 & Vanilla & 12.1 & 11.3 & 15.2 & \cellcolor[HTML]{EFEFEF}12.9 & 60.9 & 62.6 \\
 & System Prompt & 25.0 & 24.8 & 26.2 & \cellcolor[HTML]{EFEFEF}25.3 & 60.2 & 61.8 \\
 & Top-$k$ Perturbation & 42.3 & 49.4 & 49.0 & \cellcolor[HTML]{EFEFEF}46.9 & 13.3 & 8.5 \\
 & MemFree & 67.7 & 62.5 & 63.4 & \cellcolor[HTML]{EFEFEF}64.5 & 55.9 & 61.4 \\
 & R-CAD & 65.5 & 62.9 & 63.6 & \cellcolor[HTML]{EFEFEF}64.1 & 58.5 & 60.1 \\
\multirow{-6}{*}{\textbf{Gemma-2}} & \textbf{\textsc{SCoPe} (Ours)} & \textbf{74.1} & \textbf{69.7} & \textbf{71.2} & \cellcolor[HTML]{EFEFEF}\textbf{71.7} & 59.4 & 62.6 \\ \midrule
 & Vanilla & 16.5 & 14.2 & 18.3 & \cellcolor[HTML]{EFEFEF}16.3 & 59.3 & 62.5 \\
 & System Prompt & 27.2 & 26.2 & 20.8 & \cellcolor[HTML]{EFEFEF}24.7 & 59.2 & 62.5 \\
 & Top-$k$ Perturbation & 38.6 & 43.4 & 41.7 & \cellcolor[HTML]{EFEFEF}41.2 & 14.7 & 11.0 \\
 & MemFree & 60.1 & \textbf{70.6} & 62.2 & \cellcolor[HTML]{EFEFEF}64.3 & 53.5 & 60.2 \\
 & R-CAD & 68.2 & 64.5 & 68.1 & \cellcolor[HTML]{EFEFEF}66.9 & 58.8 & 61.9 \\
\multirow{-6}{*}{\textbf{Llama-3}} & \textbf{\textsc{SCoPe} (Ours)} & \textbf{73.5} & 68.6 & \textbf{68.5} & \cellcolor[HTML]{EFEFEF}\textbf{70.2} & 59.2 & 62.1 \\ \bottomrule
\end{tabular}%

\caption{Results of different methods on NewsQA. Columns 3-5 show win rates on infringement mitigation metrics, and column 6 gives the average win rate. The rightmost columns report utility preservation metrics. \textsc{SCoPe} achieves the highest average win rate in regurgitation risk while maintaining utility near the baseline.}
\label{tb:main_newsqa}
\end{table*}

\begin{table*}[!ht]
\centering
\small
\setlength{\tabcolsep}{2pt}  
\begin{tabular}{@{} 
    P{2cm}   
    P{3cm}   
    P{1.7cm}   
    P{1.7cm}   
    P{1.7cm}   
    P{1.7cm}   
    P{1.6cm}   
    P{1.6cm}   
@{}}
\toprule
 &  & \multicolumn{4}{c}{\textbf{Win Rate on Mitigation Metrics ($\uparrow$, \%)}} 
       & \multicolumn{2}{c}{\textbf{Utility Preservation ($\uparrow$)}} \\
\cmidrule(l){3-8}
\multirow{-2}{*}{\textbf{Model}} 
  & \multirow{-2}{*}{\textbf{Method}} 
  & Semantic Similarity 
  & MinHash Similarity 
  & Levenshtein Distance 
  & \cellcolor[HTML]{EFEFEF}\textbf{ Average  win rate} 
  & Blocklisted ROUGE-L 
  & In-Domain ROUGE-L \\
\midrule
& Vanilla & 5.8 & 8.5 & 6.1 & \cellcolor[HTML]{EFEFEF}6.8 & 28.1 & 32.2 \\
 & System Prompt & 30.1 & 24.8 & 29.0 & \cellcolor[HTML]{EFEFEF}28.0 & 28.1 & 31.9 \\
 & Top-$k$ Perturbation & 55.4 & 56.1 & 56.8 & \cellcolor[HTML]{EFEFEF}56.1 & 25.1 & 29.4 \\
 & MemFree & 54.2 & 55.6 & 37.1 & \cellcolor[HTML]{EFEFEF}49.0 & 28.0 & 32.1 \\
 & R-CAD & 62.5 & 61.0 & 66.5 & \cellcolor[HTML]{EFEFEF}63.3 & 28.1 & 31.5 \\
\multirow{-6}{*}{\textbf{Gemma-2}} & \textbf{\textsc{SCoPe} (Ours)} & \textbf{72.1} & \textbf{67.3} & \textbf{73.4} & \cellcolor[HTML]{EFEFEF}\textbf{70.9} & 28.1 & 32.2 \\ \midrule
 & Vanilla & 7.5 & 10.7 & 7.9 & \cellcolor[HTML]{EFEFEF}8.7 & 22.9 & 25.4 \\
 & System Prompt & 19.8 & 22.4 & 21.9 & \cellcolor[HTML]{EFEFEF}21.4 & 21.5 & 23.8 \\
 & Top-$k$ Perturbation & 58.2 & 55.1 & 61.3 & \cellcolor[HTML]{EFEFEF}58.2 & 20.6 & 22.7 \\
 & MemFree & 53.0 & 45.9 & \textbf{65.5} & \cellcolor[HTML]{EFEFEF}54.8 & 22.3 & 23.6 \\
 & R-CAD & 64.0 & 64.5 & 62.1 & \cellcolor[HTML]{EFEFEF}63.5 & 22.8 & 23.1 \\
\multirow{-6}{*}{\textbf{Llama-3}} & \textbf{\textsc{SCoPe} (Ours)} & \textbf{70.8} & \textbf{69.8} & 65.0 & \cellcolor[HTML]{EFEFEF}\textbf{68.5} & 22.3 & 23.5 \\ \bottomrule
\end{tabular}%
\caption{Results on BookSum. \textsc{SCoPe} achieves the highest average win rate while preserving utility.}
\label{tb:main_booksum}
\end{table*}
\begin{table}[t]
\centering
\small
\begin{tabular}{@{}ccc@{}}
\toprule
\textbf{Method} & \textbf{Gemma-2} & \textbf{LLama-3} \\ \midrule
Vanilla & 67.3 & 63.5 \\
System Prompt & 67.0 & 63.2 \\
Top-$k$ Perturbation & 46.1 & 45.8 \\
MemFree & 66.1 & 62.8 \\
R-CAD & 66.3 & 62.4 \\ 
\textbf{\textsc{SCoPe} (Ours)} & 66.7 & 63.1 \\ \bottomrule
\end{tabular}
\caption{MMLU accuracy across different methods with Gemma-2 and Llama-3 models.}
\label{tb:main_results_mmlu}
\end{table}

\section{Experiments}

\subsection{Setups}
\label{sec:exp_setup}

\paragraph{Datasets}
We assess copyright infringement risk using the commonly-used COTAEVAL benchmark \cite{wei2024evaluating}, which focuses on two common forms of text involved in copyrighted cases: news articles (\textsc{NewsQA}) and books (\textsc{BookSum}). Infringement is evaluated by measuring the extent to which an LLM’s continuations reproduce protected source material. The benchmark also includes blocklisted and in-domain performance retention assessments, as detailed in \citet{wei2024evaluating}. General utility is measured on the 57-task MMLU suite \cite{hendryckstest2021}. Further details are provided in Appendix~E.1.

\paragraph{Models}
All experiments use \textsc{Gemma-2-9B-it} \cite{gemma_2024} and \textsc{Llama-3-8B-Instruct} \cite{grattafiori2024llama}.  Both models have publicly released sparse autoencoders: GemmaScope \cite{lieberum2024gemmascopeopensparse} and the Llama‑3 SAE.

\paragraph{Metrics} 
Following~\cite{wei2024evaluating}, we evaluate each method in two aspects: \textbf{infringement mitigation} and \textbf{utility preservation}.
To assess copyright risk, we focus primarily on Semantic Similarity---computed via cosine similarity on embeddings produced by an off-the-shelf model---and supplement this with MinHash Similarity and Levenshtein Distance metrics for near-duplicate detection.
To ensure comparability across metrics, we adopt the \textbf{win rate} from \cite{wei2024evaluating}: the probability that a given method outperforms a randomly sampled method on a randomly selected \textsc{(metric, example)} pair, thus reflecting a method’s overall effectiveness at reducing text similarity.

For utility preservation, we follow~\citet{wei2024evaluating} and report performance on general tasks such as MMLU accuracy, as well as generation quality on \emph{in-domain} and \emph{blocklisted} prompts, to quantify any degradation in model capability. Details in Appendix~E.1.

\paragraph{Baselines}
We compare \textsc{SCoPe} against five representative methods:
(1) \textbf{\texttt{Vanilla}}: standard decoding without any protection.
(2) \textbf{\texttt{System~Prompt}}: prepends a safety-oriented instruction \cite{Anthropic2024} that discourages reproduction of copyrighted material.
(3) \textbf{\texttt{Top-$k$ Perturbation}}: adds Gaussian noise to the logits of the top-$k$ candidate tokens before sampling, to reduce memorization.
(4) \textbf{\texttt{MemFree}}~\cite{ippolito2022preventing}: filters out tokens that would form $n$-grams matching a predefined blocklist using a Bloom filter.
(5) \textbf{Reversed Context-Aware Decoding (\texttt{R-CAD})}~\cite{shi-etal-2024-trusting}: downweights token probabilities that are contextually aligned with blocklisted spans.

\subsection{Results and Observations}
\label{sec:exp_results}

Table~\ref{tb:main_newsqa} (NewsQA) and Table~\ref{tb:main_booksum} (BookSum) summarize our main experimental results. The column 3-5 report the win rate for Semantic Similarity, MinHash similarity, and Levenshtein distance, respectively, while the sixth column gives the average win rate as an overall measure of copyright-risk reduction. The final two columns and Table~\ref{tb:main_results_mmlu} show the results of Blocklisted, In-Domain, and MMLU, which quantify utility preservation.

\paragraph{Infringement mitigation performance}
Our method consistently achieves the highest risk-reduction scores on both benchmarks. Across all model and dataset combinations, \textsc{SCoPe}'s average win rate surpasses the strongest baseline (R-CAD and MemFree) by a significant margin of 3-7 percentage points, demonstrating its robust and superior suppression performance.

\paragraph{Utility preservation}
As shown in the rightmost two columns of Tables~\ref{tb:main_newsqa} and~\ref{tb:main_booksum}, as well as in Table~\ref{tb:main_results_mmlu}, our method preserves model utility while reducing infringement risk: MMLU accuracy remains unchanged, and Blocklisted and In-Domain F1 scores decline by less than 1.5 percentage points--the smallest losses among competitive baselines such as MemFree and Top-$k$ Perturbation. Although System Prompt achieves marginally better F1 preservation, its win rate is 30-40 points lower than \textsc{SCoPe}, demonstrating that sparse feature clamping provides the optimal trade-off between risk mitigation and performance retention.

Overall, the results demonstrate that \textsc{SCoPe} provides an effective, filter-free mechanism for copyright protection while fully preserving general model performance. These findings corroborate the central hypothesis that sparse, semantically aligned feature clamping can reconcile copyright-risk mitigation with task competence.

\subsection{Analysis and Discussion}

In this section, we present a targeted analysis of the subspace $\hat{\mathcal{S}}$ and our \textsc{SCoPe} intervention. We first analyze infringement mitigation performance and model utility across varying subspace sizes. Next, we apply a reverse intervention to confirm their causal effect on generation behavior. Finally, based on this experimental evidence, we discuss whether the semantic and functional properties of the copyrighted subspace align with our expectations.

\subsubsection{Impact of Subspace Size}
\label{sec:analysis_top_n}
\begin{figure}[t]
  \centering
  \includegraphics[width=\columnwidth]{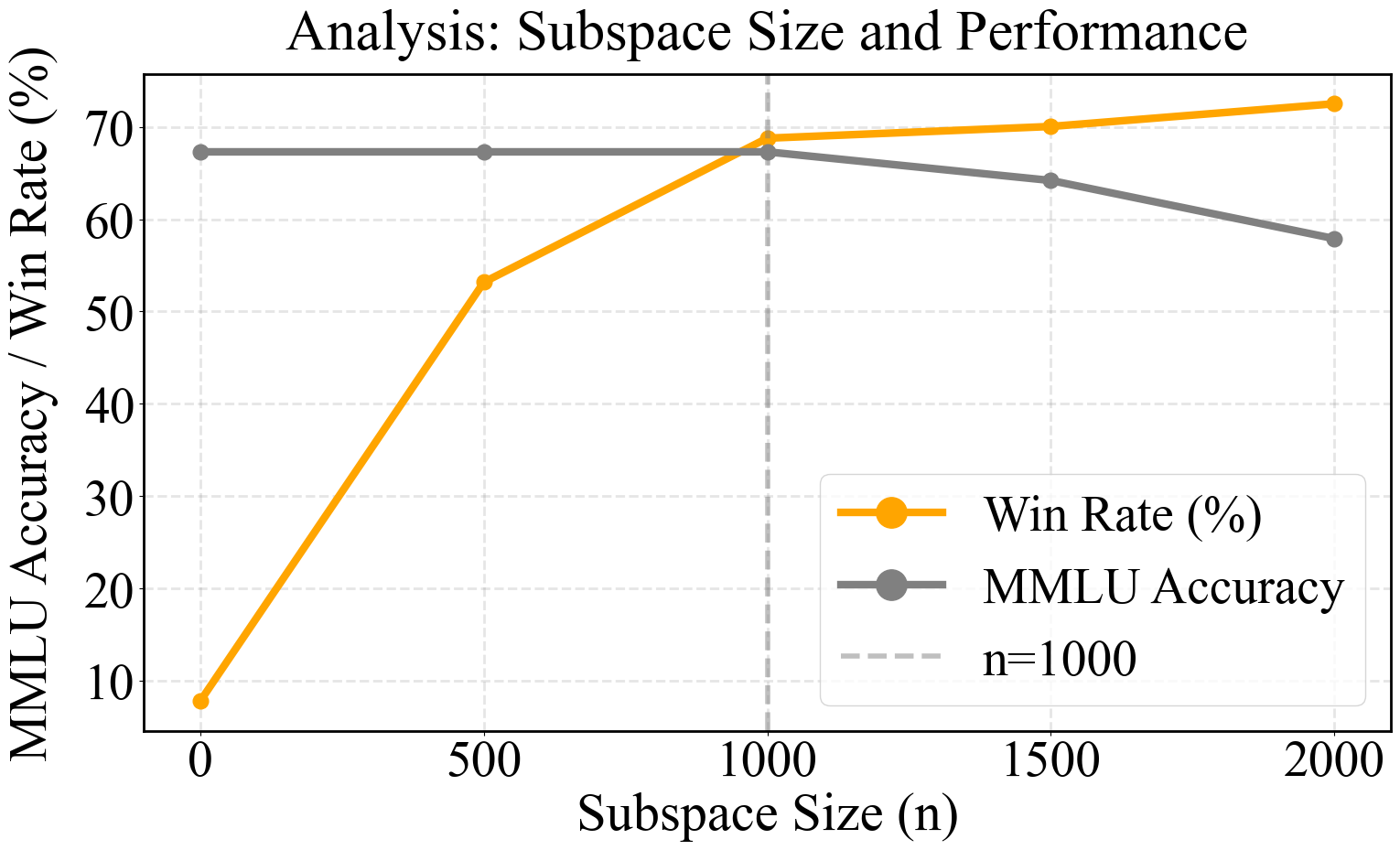}
  \caption{Analysis of the dimension \(n\) of subspace. The vertical dashed line marks the chosen setting \(n=1000\), which balances maximal risk mitigation with no loss in general utility.}
  \label{fig:ablation_top_n}
\end{figure}

Figure~\ref{fig:ablation_top_n} reports the effect of varying the dimensions $n$ of the copyrighted subspace.  As an example on the BookSum task, increasing $n$ from 0 (vanilla) to 2000, the average win rate against five baselines rises steadily (from 8.7\% to 72.5\%) indicating stronger copyright risk mitigation. However, utility remains flat only up to $n=1000$ and begins to degrade thereafter.  In particular, at $n=1000$ we achieve a win rate of 68.5\% with zero loss in MMLU, making it the optimal trade-off point.  Accordingly, we fix $n=1000$ in main experiments. Details in Appendix~E.3.

\subsubsection{Reverse Intervention: Feature Excitation}
\label{sec:reverse_intervention}

\begin{figure}[t]
  \centering
  \includegraphics[width=0.9\columnwidth]{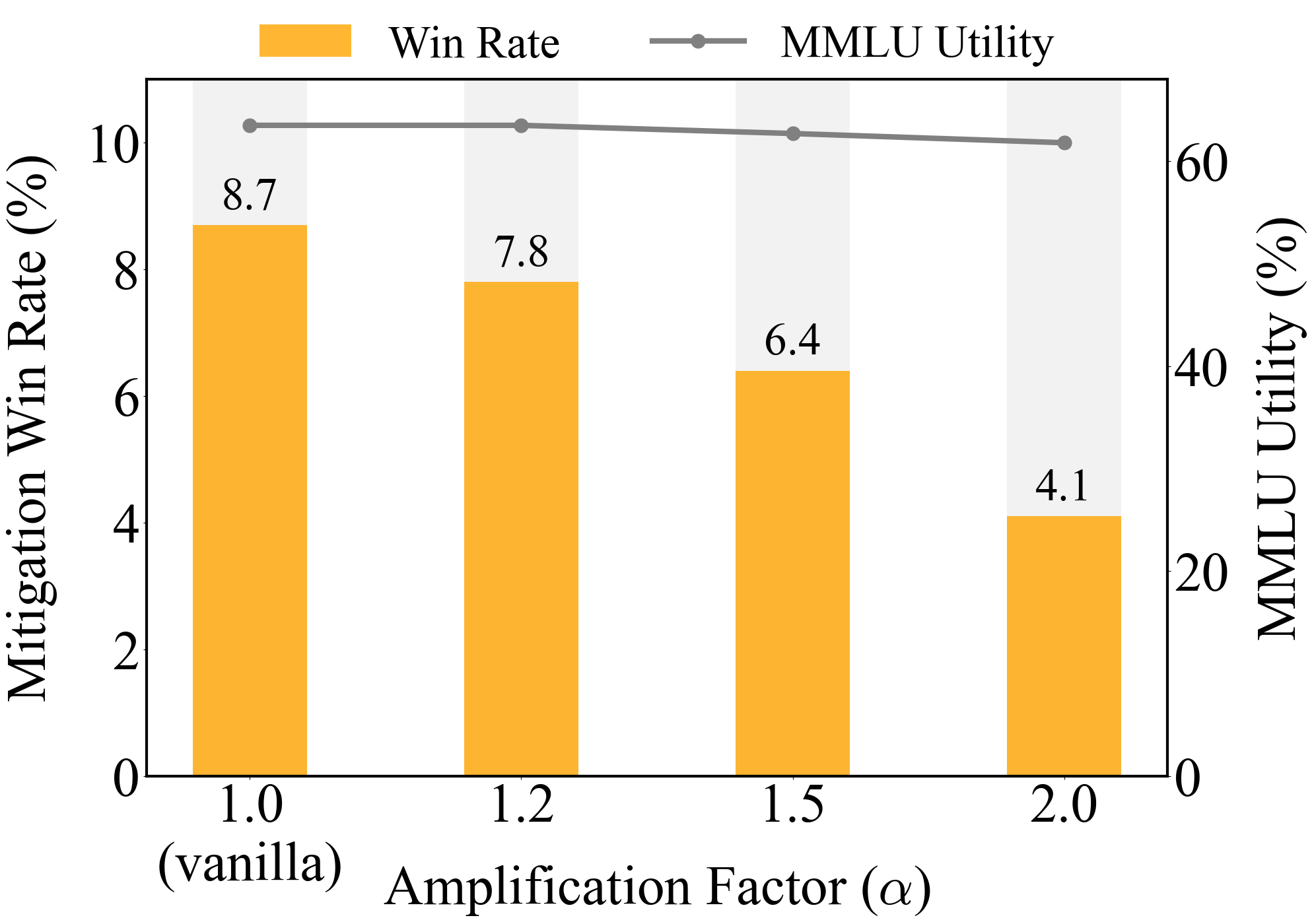}
  \caption{Impact of the reverse intervention. As we amplify the features in the copyrighted subspace $\hat{\mathcal{S}}$ with an increasing factor $\alpha$, the mitigation win rate progressively drops from 8.7\% to 4.1\%, indicating that the LLM becomes more prone to reproducing copyrighted content. This provides causal evidence that the subspace $\hat{\mathcal{S}}$ is directly responsible for generating copyrighted content.}
  \label{fig:reverse_inter}
\end{figure}

To validate the causal role of the identified subspace $\hat{\mathcal{S}}$, we conduct a reverse intervention experiment. In this setup, instead of clamping activations, we amplify the features within the copyrighted subspace at each decoding step. Specifically, for any dimension $i$ within the set of copyright dimensions $\hat{\mathcal{I}}$, we modify its activation $z_i$ by a factor of $\alpha > 1$. The modified sparse activation vector $z'$ is computed as:
\begin{equation}
z'_{i} = \begin{cases} 
\alpha \cdot z_{i}, & \text{if } i \in \hat{\mathcal{I}} \\ 
z_{i}, & \text{otherwise} 
\end{cases}
\label{eq:reverse_intervention}
\end{equation}
The results, presented in Figure~\ref{fig:reverse_inter}, show a causal link between amplifying these features and copyright infringement. We observe that as the amplification factor $\alpha$ increases from 1.0 (vanilla) to 2.0, the mitigation win rate progressively drops from 8.7\% to 4.1\%, while general utility remains largely unaffected. 
This demonstrates that our approach works bidirectionally: clamping the subspace $\hat{\mathcal{S}}$ mitigates the reproduction of copyrighted content, while conversely, amplifying it increases the LLM's propensity to generate such content. Together, these results establish a clear \textbf{causal link} between the identified subspace and the generation of copyrighted material. The observed structure and causal behavior of the subspace are consistent with our expectations.


\subsubsection{Feature Analysis and Interpretability}
\label{sec:feature_analysis}

To understand the semantic content captured by our identified subspace $\hat{\mathcal{S}}$, we performed a feature interpretability analyses. 
Our analysis reveals a clear and meaningful distinction between the \textbf{copyrighted features} (the top-n dimensions in subspace $\hat{\mathcal{S}}$ and the \textbf{general features}, which are broadly activated across both corpora (i.e., lying near the diagonal y = x  in Figure~\ref{fig:dense_sparse_space}).

Our interpretation methodology and detailed results are presented in Table~3 of Appendix~F.
Specifically, copyrighted features consistently correspond to high-level, semantic-specific concepts such as character dialogue and plot transitions. In contrast, general features relate to broader, structural patterns like formatting markers or common adjectives. This confirms that \textsc{SCoPe} operates by targeting the core semantics of protected content, not superficial stylistic or topical features.


\subsubsection{Does the Identified Copyrighted Subspace Behave as Expected?}
A critical question is whether our identified subspace $\hat{\mathcal{S}}$ genuinely captures copyright-specific semantics or merely \textit{overfits} to distributional differences between the copyrighted and general corpora.
Two key pieces of evidence argue against this overfitting hypothesis.
\begin{itemize}
\item First, our method's ability to preserve utility on unseen, in-domain data while reducing regurgitation (Tables 1 and 2) confirms that the subspace performs effective semantic isolation, rather than simply capturing topic or format features (detailed in Appendix~G.1).
\item Second, the reverse intervention experiment confirms a clear causal link, demonstrating a high degree of controllability over the generation of copyrighted content (detailed in Appendix~G.2).
\end{itemize}
Together, these experiments support that the semantic and function of the subspace $\hat{\mathcal{S}}$ is consistent with our expectations outlined in Section~\ref{sec:copyright_subspace_1}.

\section{Related Work}
\paragraph{Copyright Infringement Mitigation}
Prior work addresses LLM copyright risks at three levels.  
At the data level, corpus filtering attempts to remove licensed content before training, yet struggles with web-scale heterogeneity and often yields incomplete exclusion \cite{kandpal2022deduplicating,sag2023copyright}.  
During pretraining, methods such as Near Access-Freeness (NAF) combine auxiliary models trained only on non-copyrighted data with rejection sampling to probabilistically avoid protected passages \cite{DBLP:conf/aaai/Chu0Y24,abad2024copyright}.  Selective unlearning techniques (e.g., SUV) retroactively erase specific memorized segments but can degrade general performance \cite{xu2025suv}.  
Inference-time methods offer greater flexibility without modifying model weights. MemFree decoding avoids generating blacklisted \(n\)-grams by resampling at each step, effectively blocking known sequences but sometimes harming fluency and leaving paraphrases unchecked~\cite{ippolito2022preventing}. 
Reverse Context-Aware Decoding (R-CAD) suppresses token probabilities linked to protected content, yielding strong leakage reduction at the cost of increased computation and potential over-suppression~\cite{shi-etal-2024-trusting}. These methods mostly rely on external corpora or Bloom filter indexes and are effective only for exact verbatim matches, and may lead to hallucinations~\cite{liu2024shield}.  
These methods mostly rely on external corpora or Bloom filter indexes and are effective only for exact verbatim matches, and may lead to hallucinations~\cite{liu2024shield}.

\paragraph{Activation Steering by SAEs}  
Activation steering methods adjust neuron activations at inference time to guide LLM behavior without fine-tuning. Sparse autoencoders (SAEs) learn high-dimensional, semantically disentangled feature spaces from model internals, providing interpretable axes for control \cite{gao2024scaling}. SAE-Targeted Steering (SAE-TS) selects steering vectors that specifically amplify or suppress chosen SAE features, improving precision over vanilla activation additions \cite{chalnev2024improving}. Feature Guided Activation Additions (FGAA) further refine this approach by optimizing steering vectors in the SAE latent space, yielding stronger and more coherent steering effects across diverse tasks \cite{soo2025steering}. A related mechanism, feature clamping, thresholds SAE activations to suppress unwanted concepts in real time \cite{bricken2023monosemanticity}. In summary, SAE techniques offer a lightweight and transparent way to steer LLM behavior.

\section{Conclusion}
In this work, we cast copyright infringement mitigation for LLMs as intrinsic semantic space control and introduce \textsc{SCoPe}, an inference-time method that isolates a copyright-sensitive subspace and suppresses its influence during decoding.
Experimental results demonstrate that it significantly reduces the reproduction of copyrighted content while maintaining overall generation quality. We also provide experimental evidence to validate that the identified subspace is functionally specific to copyrighted text. Further interpretability analyses corroborate this finding, confirming that the suppressed dimensions encode high-level semantics rather than surface patterns.
However, our approach has limitations. It is currently restricted to open-source models with publicly available SAEs, as it requires access to their intermediate hidden states. Additionally, modeling the copyright region as a linear subspace is an approximation, suggesting future work on adaptive or non-linear methods.

\section*{Ethical Statement}
In this research, we use only publicly available models and datasets and collect no personal or sensitive data. All evaluations rely on openly licensed benchmarks, and no human subjects were involved.
\textsc{SCoPe} aims to reduce copyright infringement risk and bolster model reliability. No actual copyrighted material was used; benchmarks simulate protected content using publicly available data.
AI tools were used only for text polishing, not for research design or analysis.

\section*{Acknowledgements}
This work was supported by Beijing Natural Science Foundation (L253001) and Key Laboratory of Science, Technology and Standard in Press Industry (Key Laboratory of Intelligent Press Media Technology). We appreciate the anonymous reviewers for their helpful comments. Xiaojun Wan is the corresponding author.

\bibliography{aaai2026}
\newcommand{\isChecklistMainFile}{}
\newpage
\appendix
\section{SAE-Induced Sparse Semantic Space}
\label{app:sae_details}

To ensure that the SAE produces disentangled and monosemantic activations, we enforce sparsity at both the encoding and training stages.

\subsection{JumpReLU Activation}  
JumpReLU extends the standard ReLU by introducing a positive threshold \(\tau>0\), zeroing out any pre-activation below \(\tau\). Formally, for each scalar \(x\):
\begin{align}
\mathrm{JumpReLU}_\tau(x) \;=\; x \,\cdot\, H(x - \tau)
\label{eq:jumprelu}
\end{align}

where \(H(\cdot)\) is the Heaviside step function. By “jumping” at \(\tau\), only activations exceeding the threshold are retained, which fosters sparser, more disentangled representations while preserving the magnitude of salient features.

\subsection{Sparsity Regularization During Training}
During SAE training, we combine reconstruction fidelity with an \(\ell_{1}\) penalty on the sparse code \(\mathbf{z}\).  Given an input hidden state \(\mathbf{h}\) and its reconstruction \(\hat{\mathbf{h}}\), the loss is:
\[
\mathcal{L}_{\mathrm{SAE}}(\mathbf{h})
= \|\hat{\mathbf{h}} - \mathbf{h}\|_{2}^{2}
+ \lambda\,\|\mathbf{z}\|_{1},
\]
where \(\mathbf{z}=f(\mathbf{h})\) is the JumpReLU output, \(\|\mathbf{z}\|_{1}\) its \(\ell_1\) norm, and \(\lambda>0\) balances reconstruction accuracy against sparsity.  This regularization drives most dimensions of \(\mathbf{z}\) to zero, activating only those channels that carry semantically salient information, and thus yielding an interpretable sparse semantic space.

\section{Copyright Alignment Score}
\label{app:score}

\subsection{Interpretation of \(\mathcal{Q}\)}
\label{app:meaning_of_score}

\paragraph{Subspace score.}
For a candidate subspace \(\mathcal{S}\), $\mathcal{Q}(\mathcal{S})$ is the probability that, after \textit{independently} drawing
(i) a feature $i$ from the index set of $\mathcal{S}$,
(ii) a copyrighted sample $x_{\mathrm{cr}}\!\sim\!\mathcal{C}_{\mathrm{cr}}$, and
(iii) a general sample $x_{\mathrm{gen}}\!\sim\!\mathcal{C}_{\mathrm{gen}}$,
the activation of that feature on $x_{\mathrm{cr}}$ exceeds its activation on $x_{\mathrm{gen}}$.
Values near $1$ signal both \emph{coverage} (features fire on most copyrighted texts) and \emph{exclusivity} (they rarely fire on general texts); a value of $0.5$ corresponds to random chance.

\paragraph{Single-feature score.}
In practice, we compute the alignment score for each dimension individually. The per-dimension variant is:
\begin{align}
     \mathcal{Q}(i) 
= 
\mathbb{E}_{\substack{x_{\mathrm{cr}} \sim \mathcal{C}_{\mathrm{cr}} \\ x_{\mathrm{gen}} \sim \mathcal{C}_{\mathrm{gen}}}}
\left[ 
  \mathbb{I}\left( z_i^{(x_{\mathrm{cr}})} > z_i^{(x_{\mathrm{gen}})} \right) 
\right]
\end{align}
This score characterizes the alignment between dimension $i$ and the ideal copyrighted subspace $\mathcal{S}^*$. Formally, $\mathcal{Q}(i)$ approximates the area under the ROC curve (AUROC) for a binary classifier that discriminates $\mathcal{C}_{\mathrm{cr}}$ (copyrighted) and $\mathcal{C}_{\mathrm{gen}}$ (general) using $z_i^{(x)}$ as the decision score. We rank all features by \(\mathcal{Q}(i)\) and keep the
top \(n\): this greedy rule is linear-time and empirically yields a
high-quality copyrighted subspace.

\subsection{Bounding Property}
\label{app:bound}

Recall the definition of the copyright-alignment score for a subspace
\(\mathcal{S}\) whose index set is \(\mathcal{I}\):
\[
\mathcal{Q}(\mathcal{S})
=\underset{\substack{i\sim\mathcal{I}\\
x_{\mathrm{cr}}\sim\mathcal{C}_{\mathrm{cr}}\\
x_{\mathrm{gen}}\sim\mathcal{C}_{\mathrm{gen}}}}
{\mathbb{E}}
\!\Bigl[
\mathbb{I}\!\bigl(z_i^{(x_{\mathrm{cr}})}
              > z_i^{(x_{\mathrm{gen}})}\bigr)
\Bigr],
\]
where \(i\) is drawn uniformly from \(\mathcal{I}\).  
For clarity, define the \emph{per-feature} score
\(
\mathcal{Q}(i)
:=\Pr\!\bigl[z_i^{(x_{\mathrm{cr}})}
             >z_i^{(x_{\mathrm{gen}})}\bigr].
\)
Because \(i\) is chosen uniformly, the subspace score is simply the
average of these per-feature scores:
\[
\mathcal{Q}(\mathcal{S})
=\frac{1}{|\mathcal{I}|}\sum_{i\in\mathcal{I}}\mathcal{Q}(i).
\]

Every term \(\mathcal{Q}(i)\) is bounded above by the largest
per-feature score in the same set, denoted
\(\max_{j\in\mathcal{I}}\mathcal{Q}(j)\).  Replacing each summand with
this maximum yields an upper bound on the average:
\[
\mathcal{Q}(\mathcal{S})
\le
\frac{1}{|\mathcal{I}|}
\sum_{i\in\mathcal{I}}
\max_{j\in\mathcal{I}}\mathcal{Q}(j)
=
\max_{j\in\mathcal{I}}\mathcal{Q}(j).
\]

\noindent Hence we obtain:
\[
\boxed{
\mathcal{Q}(\mathcal{S})
\le
\max_{i\in\mathcal{I}}\mathcal{Q}(i)
}.
\]

\smallskip
\textbf{Interpretation.}  
The inequality states that a subspace cannot score higher than its
best individual feature.  Adding features with lower alignment can
only dilute (or leave unchanged) the overall score, so subspace
construction should focus on identifying high-scoring dimensions
rather than aggregating many mediocre ones.

\section{Details of Empirical Validation Study}
\label{app:em_study}

\subsection{Settings of Empirical Study}

In our empirical study, the \textbf{copyrighted corpus} $\mathcal{C}_{cr}$ is the NewsQA subset from the COTAEVAL benchmark~\cite{wei2024evaluating}. 

The \textbf{general corpus} $\mathcal{C}_{gen}$ used in this experiment consists of 15 k examples uniformly sampled across categories from MMLU.

We use the \textsc{Gemma-2-9b-it} model and attach a pretrained sparse autoencoder (SAE) at the 20th residual stream. The SAE projects each hidden state \(\mathbf{h}\) into a sparse encoding \(\mathbf{z}=[z_1,\dots,z_k]\).

For each feature \(z_i\), we compute:
\begin{enumerate}
  \item The maximum activation \(\max_t z_i(t)\) over all timesteps \(t\) for each input sample.
  \item The average of these maximum activations across samples in the \emph{copyrighted} subset.
  \item The average of these maximum activations across samples in the \emph{general} subset.
\end{enumerate}

We then plot these two averages for every feature in a two-dimensional scatter plot (Figure~1a). In this figure:
\begin{itemize}
  \item The dense hidden state \(\mathbf{h}\) activations cluster tightly around the diagonal line \(y = x\), indicating similar response to both copyrighted and general texts.
  \item The sparse SAE features exhibit a broader distribution, with several features concentrated in the upper-left region. These features activate strongly on copyrighted inputs but weakly on general inputs, suggesting a distinct copyright-sensitive subspace.
\end{itemize}

This visualization empirically supports the existence of a semantically coherent copyrighted subspace and motivates our subsequent algorithm for its estimation.  

\subsection{Activation Threshold $\tau$}

The JumpReLU activation in our SAE (see Equation~\ref{eq:jumprelu}) uses a positive threshold \(\tau\), so that each feature \(z_i\) only becomes active when its pre-activation exceeds \(\tau\). To determine a suitable \(\tau\), we conducted an empirical analysis of the maximum activation values across all dimensions. Figures~1b and 1c plot these maxima and reveal a natural separation at approximately \(\tau=5\). In practice, all pre-activations below 5 are zeroed out, while those above 5 pass through unchanged. This choice balances sparsity and signal preservation in the semantic space.  

\subsection{Additional Experimental Results}
\label{app:additional_em_results}

\begin{figure*}[t]
  \centering
  \begin{subfigure}[t]{0.48\textwidth}
    \includegraphics[width=\textwidth]{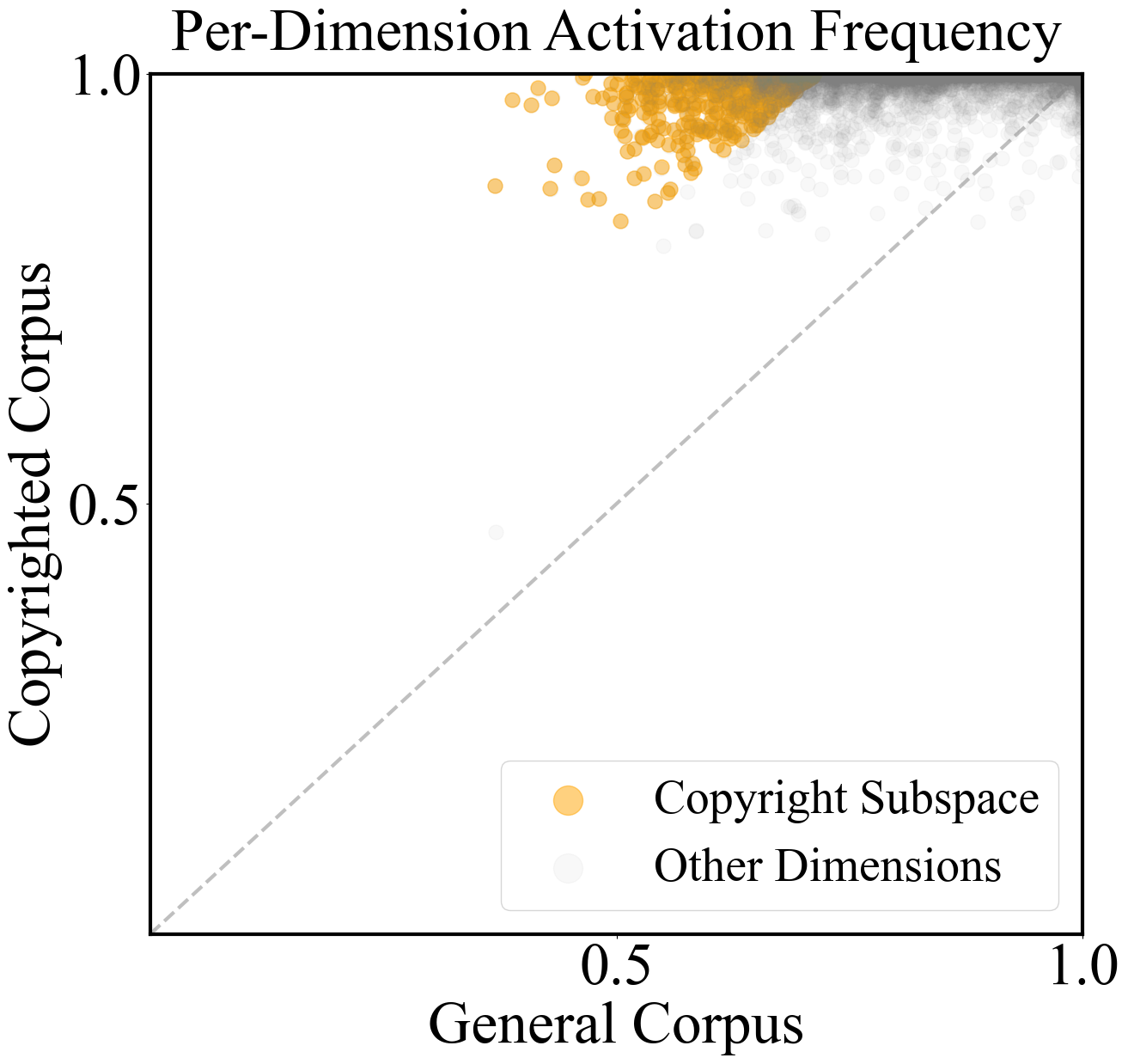}
    \caption{Dense LLM feature activation frequencies}
    \label{fig:act_freq_dense}
  \end{subfigure}
  \hfill
  \begin{subfigure}[t]{0.48\textwidth}
    \includegraphics[width=\textwidth]{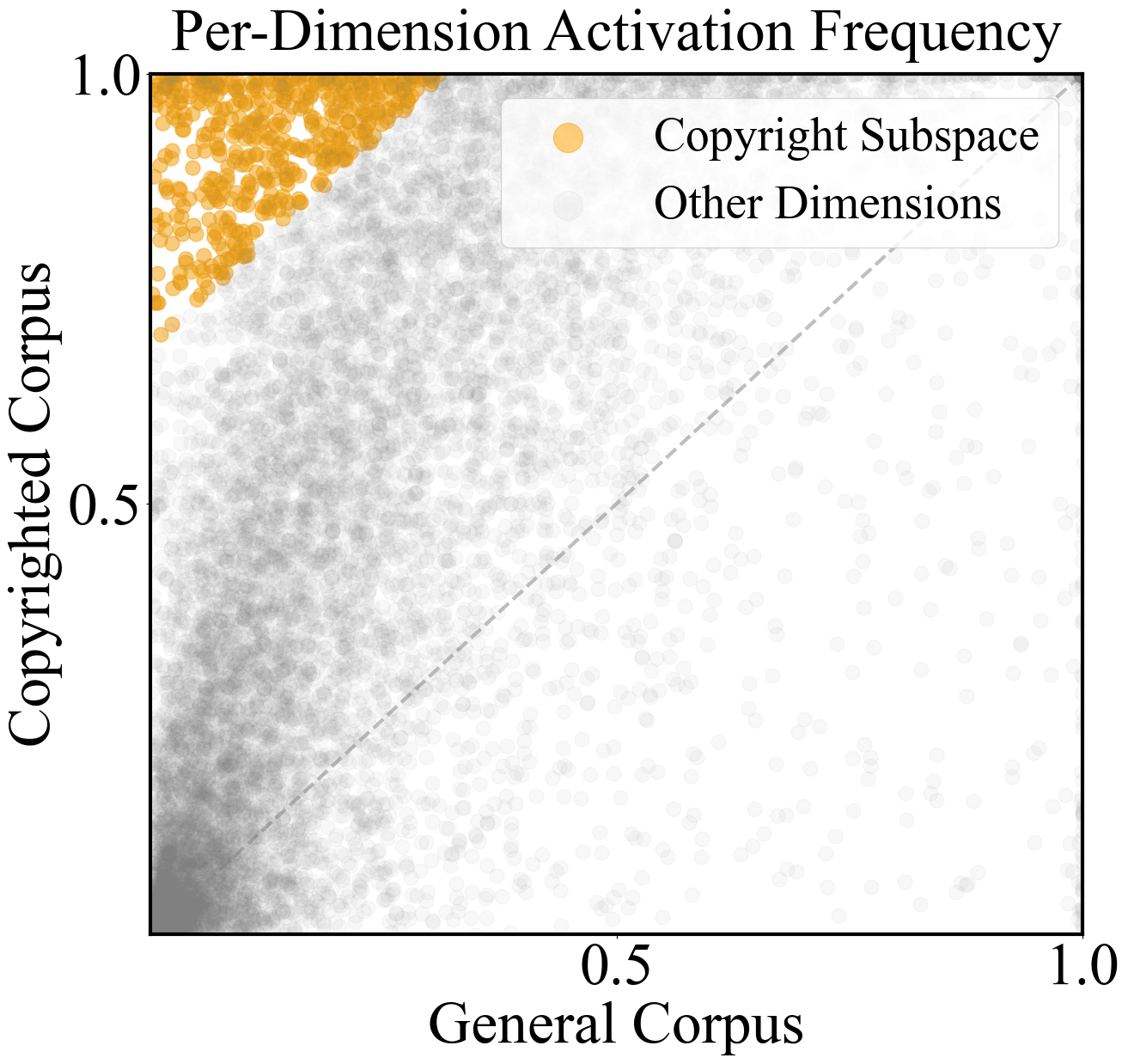}
    \caption{Sparse SAE feature activation frequencies (copyrighted subspace in orange)}
    \label{fig:act_freq_sparse}
  \end{subfigure}
  \caption{Activation frequency distributions for (a) dense LLM features, which cluster tightly in the upper‐right with no clear separation of copyright‐sensitive dimensions, and (b) sparse SAE features, where the copyrighted subspace (orange) forms a distinct cluster in the upper‐left region.}
  \label{fig:activation_frequencies}
\end{figure*}

We further analyze the activation frequency of each feature by computing, for every sample, the maximum activation value of each dimension and marking a feature as “active” if this value exceeds the threshold \(\tau\). We then calculate the activation rate across all samples, yielding a value in \([0,1]\). Figures~\ref{fig:act_freq_dense} and~\ref{fig:act_freq_sparse} show the resulting distributions for the dense LLM representations and the sparse SAE features, respectively.  

Using the method described in Section~3.1, we construct a copyright-sensitive subspace of size 500. In Figure~\ref{fig:act_freq_sparse}, the features belonging to this subspace are highlighted in orange.  

\paragraph{Analysis of Activation Frequencies}
In the dense semantic space (Figure~\ref{fig:act_freq_dense}), all feature activations cluster tightly in the upper-right corner, indicating that every dimension fires frequently on both general and copyrighted texts. Even when our subspace construction method is applied here, the resulting features remain mixed with the overall representation and exhibit no clear separation. By contrast, in the sparse semantic space (Figure~\ref{fig:act_freq_sparse}), a distinct group of dimensions appears in the upper-left region. These dimensions, highlighted in orange, activate strongly on copyrighted inputs while remaining largely silent on general text, thereby forming a clearly separable copyright-sensitive subspace. This contrast demonstrates that sparse encoding is effective at isolating high-level semantic features relevant to infringement risk.

\begin{algorithm}[h]
\caption{Estimating the Empirical Copyrighted Subspace \( \hat{\mathcal{S}} \)}
\label{alg:subspace_estimation}
\DontPrintSemicolon
\KwIn{
  Copyrighted corpus \( \mathcal{C}_{\mathrm{cr}} \), general corpus \( \mathcal{C}_{\mathrm{gen}} \)\\
  SAE encoder \( f: \mathbb{R}^d \rightarrow \mathbb{R}^k \)\\
  Activation threshold \( \tau \in \mathbb{R} \)\\
  Number of selected dimensions \( n \)
}
\KwOut{Estimated subspace \( \hat{\mathcal{S}} \subset \mathbb{R}^k \)}

\vspace{0.5em}
\textbf{Step 1: Extract sparse activations from both corpora} \\
\Indp
\ForEach{sample \( x \in \mathcal{C}_{\mathrm{cr}} \cup \mathcal{C}_{\mathrm{gen}} \)}{
    Extract hidden state \( \mathbf{h}^{(x)} \) from frozen LLM \\
    \( \mathbf{z}^{(x)} \leftarrow f(\mathbf{h}^{(x)}) \)
}
\Indm

\vspace{0.5em}
\textbf{Step 2: Compute per-dimension scores} \\
\Indp
\ForEach{dimension \( i = 1, \dots, k \)}{
    Define binary indicator \( a_i^{(x)} \leftarrow \mathbb{I}(z_i^{(x)} > \tau) \) \\
    Compute alignment score \( \mathcal{Q}(i) \) distinguishing \(\mathcal{C}_{\mathrm{cr}}\) vs. \(\mathcal{C}_{\mathrm{gen}}\)
}
\Indm

\vspace{0.5em}
\textbf{Step 3: Select top-$n$ scoring dimensions} \\
\Indp
Sort all dimensions \( i \) by score \( \mathcal{Q}(i) \) in descending order \\
Let \( \hat{\mathcal{I}} \leftarrow \) indices of top-$n$ dimensions \\
Let \( \hat{\mathcal{S}} \leftarrow \mathcal{S}(\hat{\mathcal{I}}) \)
\Indm

\vspace{0.5em}
\KwRet \( \hat{\mathcal{S}} \)
\end{algorithm}

\section{\textsc{SCoPe} Algorithm Details}
\label{app:scope}

\paragraph{Algorithm for Estimating the Copyrighted Subspace}  
Algorithm~\ref{alg:subspace_estimation} outlines our procedure for identifying the empirical copyrighted subspace \(\hat{\mathcal{S}}\). In Step 1, we extract sparse activation vectors \(\mathbf{z}^{(x)}\) from both the copyrighted corpus \(\mathcal{C}_{\mathrm{cr}}\) and a general corpus \(\mathcal{C}_{\mathrm{gen}}\) by passing each sample \(x\) through the frozen LLM and its pretrained SAE encoder. Step 2 converts these real-valued activations into binary signals via thresholding at \(\tau\), and computes a per-dimension alignment score \(\mathcal{Q}(i)\) that measures how consistently each feature \(i\) fires more strongly on copyrighted text than on general text. In Step 3, we rank dimensions by \(\mathcal{Q}(i)\) and select the top-\(n\) indices \(\hat{\mathcal{I}}\), thereby defining the subspace \(\hat{\mathcal{S}} = \{\mathbf{z}\mid z_i=0\ \forall\,i\notin\hat{\mathcal{I}}\}\). This three-stage process yields a compact, interpretable subspace whose features are highly predictive of copyright-sensitive content.

\section{Main Experiment Supplements}
\label{app:exp}

\subsection{Experimental Setup}
\label{app:exp_setup}

\paragraph{Sparse autoencoders}
Owing to compute limits, we employ publicly released SAEs rather than training from scratch.
\begin{itemize}
    \item For \textbf{Gemma-2-9B-IT} we attach the SAE trained on the 20th residual stream (\(\ell=20\)), which employs a JumpReLU activation with threshold \(\tau=5\).
    \item For \textbf{Llama-3-8B-Instruct} we use the SAE on the 25th residual stream (\(\ell=25\)); this model adopts a gated activation rather than JumpReLU, but keeps the same 16 k dictionary size.
\end{itemize}

\paragraph{Hardware and Runtime}
All experiments were conducted on an internal cluster of NVIDIA A40 GPUs (48 GB each) running Ubuntu 20.04 LTS with CUDA 12.0. End-to-end evaluation across all model–dataset pairs required approximately 80–100 GPU hours.

\paragraph{Datasets and Evaluation}  
Our experiments follow the \textsc{COTAEVAL} protocol but extend the original splits to ensure balanced coverage. Each instance consists of:  
\begin{itemize}
  \item A source document of up to 400 tokens.  
  \item A prompt for autoregressive completion (\texttt{prompt\_autocomplete}), taken as the first 200 tokens.  
  \item A ground-truth continuation (\texttt{gt\_autocomplete}), taken as tokens 201–400.  
\end{itemize}
To assess infringement risk, we feed \texttt{prompt\_autocomplete} into each LLM and compare its generated continuation against \texttt{gt\_autocomplete} using three metrics: Semantic Similarity, MinHash Similarity, and Levenshtein Distance. We then pool all \((\text{sample},\ \text{metric})\) pairs across methods, randomly draw comparisons, and compute the win rate of each method as the fraction of pairwise wins over all baselines.

\paragraph{Utility Evaluation}  
We assess general capability using accuracy on the 57-task \textbf{MMLU} benchmark~\cite{hendryckstest2021}. To measure downstream performance with minimal copyright risk, we evaluate each model on two held-out sets for the target task (BookSum summarization or NewsQA document QA):  
\begin{itemize}
  \item \emph{Blocklisted set}: the examples drawn from the copyrighted corpus used during subspace estimation, testing whether models retain task performance on familiar (but copyright-sensitive) data.  
  \item \emph{In-domain set}: new examples of the same task drawn from non‐copyrighted but task‐matched data, assessing performance on unseen inputs of the same type.  
\end{itemize}
For BookSum, we report ROUGE scores; for NewsQA, we report QA F1. Comparing blocklisted and in-domain results ensures that \textsc{SCoPe} protects against copying while preserving utility both on data similar to the protected corpus and on entirely new in-domain examples.  

During evaluation, all LLMs are decoded with greedy sampling ( \texttt{do\_sample = False} ).

\subsection{Baselines}
\paragraph{Vanilla LLM (No Protection)}  
The base model is used without any intervention, employing greedy decoding to generate continuations. 

\paragraph{System Prompt}  
A meta-instruction is prepended to the input to discourage copying, for example:  
\begin{tcolorbox}[promptbox]
You are a helpful, respectful and honest assistant. When generating your response, please do not generate verbatim reiteration of the information put in your context...
\end{tcolorbox}
This method relies on the model’s internal compliance with the instruction to reduce unintended verbatim reproduction. 

\paragraph{Top-\(k\) Perturbation}  
At each sampling step, Gaussian noise is added to the logits of the top-\(k\) highest-probability tokens before selecting the next token. This perturbation aims to disrupt memorized sequences and lower exact copying, though excessive noise can impact fluency and factual correctness. 
\paragraph{MemFree Decoding}  
A Bloom-filter–based blocklist of \(n\)-grams from copyrighted passages is maintained. During decoding, any candidate token that would complete a blocklisted \(n\)-gram is discarded, and the next highest-scoring token is considered. This filter effectively prevents verbatim repeats but may be bypassed by paraphrases or minor edits. 
\paragraph{Reversed Context-Aware Decoding (R-CAD)}
At each decoding step, R-CAD adjusts token probabilities by down-weighting contributions from retrieved blocklisted content. Specifically, it subtracts the logits of candidate tokens conditioned on the blocklisted context from those conditioned on the user query. This reduces the influence of memorized protected content during generation. While effective at suppressing regurgitation, R-CAD is limited to memorization scenarios and incurs significant computational overhead due to repeated context-based scoring.

\subsection{Analysis of Subspace Size}
\label{app:ablation}

\begin{table}[t]
\centering
\small
\setlength{\tabcolsep}{1pt}
\begin{tabular}{@{} 
    P{1.4cm}    
    P{2cm}    
   P{1cm}    
    P{1.5cm}    
    P{1.5cm}    
@{}}
\toprule
 & \textbf{\begin{tabular}[c]{@{}c@{}}Risk Reduction\end{tabular}}
   & \multicolumn{3}{c}{\textbf{Utility ($\uparrow$)}} \\
\cmidrule(l){3-5}
\multirow{-2}{*}{\textbf{\begin{tabular}[c]{@{}c@{}}Setting\\ top-$n$\end{tabular}}}
  & \cellcolor[HTML]{EFEFEF}\textbf{win rate (\%,~$\uparrow$)}
  & MMLU
  & \begin{tabular}[c]{@{}c@{}}Blocklisted\\ ROUGE-L\end{tabular}
  & \begin{tabular}[c]{@{}c@{}}In-Domain\\ ROUGE-L\end{tabular} \\
\midrule
$0$(Vanilla) & \cellcolor[HTML]{EFEFEF}8.7 & 63.5 & 22.9 & 25.4 \\
$500$ & \cellcolor[HTML]{EFEFEF}53.2 & 63.5 & 22.9 & 24.8 \\
{$\mathbf{1000}$} & \cellcolor[HTML]{EFEFEF}\textbf{68.5} & \textbf{63.1} & 22.3 & 23.5 \\
$1500$ & \cellcolor[HTML]{EFEFEF}70.1 & 61.8 & 20.5 & 20.7 \\
$2000$ & \cellcolor[HTML]{EFEFEF}72.5 & 57.9 & 17.6 & 18.4 \\ \hline
\end{tabular}
\caption{Analysis over the dimension $n$ of subspace on \textsc{Gemma-2-9B-it}.  
  While increasing $n$ steadily improves risk reduction (win rate), utility begins to degrade beyond $n=1000$.  
  Setting $n=1000$ achieves the best balance, delivering high win rate with no loss in MMLU.} 
\label{tb:ablation_top_n}
\end{table}

In Table~\ref{tb:ablation_top_n}, each win rate for a given subspace size \(n\) is calculated by averaging \textsc{SCoPe}’s performance at that \(n\) against the four external baselines. This facilitates direct comparison across settings, rather than conducting pairwise comparisons between different \(n\) values. The win rate at \(n=0\) (the vanilla LLM) is taken directly from Table~2 in the main text.

\subsection{Reverse Intervention}
\label{app:reverse_intervention}

Table~\ref{tb:reverse_intervention} reports the win rate on BookSum for $\alpha\in\{1.2,1.5,2.0\}$.  We implement activation amplification by multiplying each feature in $\hat{\mathcal{S}}$ by $\alpha$ immediately before the $W_{dec}$ projection.

\begin{table}[h]
\small
\centering
\begin{tabular}{@{}ccc@{}}
\toprule
\textbf{Setting} & \textbf{Mitigation} & \textbf{Utility} \\
$\alpha$ & \textbf{win rate (\%,$\uparrow$)} & \textbf{MMLU($\uparrow$)} \\ \midrule
$1$ (vanilla) & 8.7 & 63.5 \\
$1.2$ & 7.8 & 63.5 \\
$1.5$ & 6.4 & 62.7 \\
$2.0$ & 4.1 & 61.8 \\ \bottomrule
\end{tabular}
\caption{Results of reverse feature activation on BookSum. }
\label{tb:reverse_intervention}
\end{table}

Table~\ref{tb:reverse_intervention} reports how scaling the clamped dimensions by $\alpha>1$ affects both risk and utility on the BookSum task.  When $\alpha$ increases from 1.0 to 2.0, the win‐rate against the four baselines falls from 8.7\% to 4.1\%, indicating that stronger feature activation reinstates more copyrighted content.  MMLU accuracy remains effectively unchanged for amplification ($\alpha$).  

These results demonstrate a clear causal link between the selected subspace dimensions and narrative content generation. As we amplify the clamped features, the model increasingly reproduces copyrighted passages, showing that these dimensions directly encode thematic and plot‐related signals. Moderate amplification reinstates specific story elements without disrupting overall reasoning, while excessive amplification overrides general competence and floods the output with memorized text.

\section{Feature Analysis Supplement}
\label{app:feature_analysis_supplement}
In interpretability studies, each dimension of the sparse space is commonly referred to as a \textbf{feature}.
To understand the semantic meaning of each feature (i.e., each dimension in the sparse space), we employ automated interpretability tools. Specifically, we use the \textbf{Auto-Interp} method, where Neuronpedia provides numerous top activation snippets for a given feature to a powerful LLM (e.g., GPT-4o-mini). The LLM then synthesizes this information to generate a concise, human-readable description of the feature’s function; see~\cite{neuronpedia} for details.

To further analyze a feature's influence on token generation, we also utilize the \textbf{Logit Lens} analysis. This technique projects a feature's direction through the model's final unembedding layer to reveal which vocabulary tokens are most strongly promoted (top positive logits) or suppressed (top negative logits) by its activation.

\subsection{Additional Activation Examples}  

To evaluate the interpretability and semantic specificity of our clamped subspace, we analyze two categories of features (i.e., dimensions) on the BookSum task: (1) \textbf{general features} activate broadly across corpora (lie on the $y=x$ in the Figure~1a Right), and (2) \textbf{copyrighted features}, which are the top-3 dimensions from our identified subspace, selected according to \textit{Copyright Alignment Score} $\mathcal{Q}(i)$.

Table~\ref{tb:feature_analysis_supplement} provides extra illustrative cases, showing the activation values of selected features (i.e., dimensions) at each token. 

Overall, a clear semantic difference exists between the two categories. General features correspond to broad, structural patterns, such as common descriptive adjectives (e.g., \texttt{great}, \texttt{grand}), formatting markers, and analytical phrases. In contrast, copyrighted features capture semantic attributes more specific to the protected corpus, showing a distinct difference from the general features.

\begin{table*}[t]
\centering
 \small
\begin{tabularx}{\textwidth}{@{}  >{\centering\arraybackslash}p{1cm}  >{\centering\arraybackslash}p{1cm}  p{2cm} >{}p{10cm} @{}}
\toprule
\textbf{Subspace} & \textbf{Feature ID} & \textbf{Interpretation} & {\small\textbf{Activation Example}} \\
\midrule
\multirow{3}{*}{\textbf{General}}
  & 755   & Particularly relating to “great” and “grand”.   
          & \makecell[l]{%
         \texttt{1.~My \hlorange{great}{13.156}-\hlorange{great}{35.110} uncle was a very young boy at the time...} \\
         \texttt{2.~...any Parents,\hlorange{~grandparents}{11} or\hlorange{~great}{35}-\hlorange{grand}{19}\hlorange{parents}{7} of such} \\
         \texttt{~~~a person, on the male or female line...}\\
         \texttt{3.~Hattie Lee, died in 1893 from a tragic fire) amaternal} \\
         \texttt{~~~\hlorange{great}{22}\hlorange{~great}{36} grandmother.}
         } \\ \addlinespace[0.5em]
  & 10184 & Occurrences of bold text formatting.  
        & \makecell[l]{
        \texttt{1.~...as many as you want on the \hlorange{<b>}{42.591}Actions</b> page...} \\
        \texttt{2.~The \hlorange{<b>}{48}Appearance</b> options can be set to avoid fading the} \\
        \texttt{~~~screen via the \hlorange{<b>}{50}None</b> theme...}\\
        \texttt{3.~you can use the \hlorange{<b>}{46}Check for Updates</b> feature in the app} \\
        \texttt{~~~to update.} 
        } \\ \addlinespace[0.5em]
  & 11848 & Analyses or evaluations of data and findings.  
        & \makecell[l]{
        \texttt{1.~<bos> the\hlorange{~analysis}{93}\hlorange{~of}{57}\hlorange{~the}{24} NMR data\hlorange{~from}{8}\hlorange{~a}{5} range of } \\
        \texttt{~~~oil-in-water emulsions\hlorange{~are}{12} evaluated...}\\
        \texttt{2.~\#\#\#\hlorange{~Analysis}{95}\hlorange{~of}{60} primary objectives \{\#Sec17\}} \\
        \texttt{3.~However, this\hlorange{~analysis}{94} necessarily in\hlorange{volves}{73.98} deciding whether} \\
        \texttt{~~~the district court correctly applied Maryland law...}
        } \\ \addlinespace[0.5em]
\midrule
\multirow{3}{*}{\textbf{Copyright}}
  & 15445 & Dialogue and interactions between characters.  
        & \makecell[l]{
        \texttt{1.~\hlorange{“}{62}\hlorange{You}{19}\hlorange{~will}{20}\hlorange{~stand}{5}\hlorange{~up}{5}\hlorange{,”}{14} the Queen\hlorange{~ordered}{15} ... \hlorange{“}{69}\hlorange{You}{24}\hlorange{~will}{19}\hlorange{~answer}{8}} \\
        \texttt{~~~\hlorange{~the}{8}\hlorange{~question}{7}\hlorange{.”}{5.6}} \\
        \texttt{2.~\hlorange{"}{70}\hlorange{They}{21}\hlorange{~are}{28}\hlorange{~all}{10} waiting for us in the Great Hall\hlorange{,"}{13} Gabrielle } \\
        \texttt{~~~\hlorange{~said}{32}. \hlorange{"}{47}\hlorange{Let}{11} them wait\hlorange{,"}{10} the Conqueror\hlorange{~smiled}{13} and looked} \\
        \texttt{~~~\hlorange{~at~her}{8} Queen...}\\
        \texttt{3.~\hlorange{“}{64}\hlorange{Evening}{11}\hlorange{,}{31}\hlorange{~Miss}{8} McKinley\hlorange{.”}{16} \hlorange{“}{42}\hlorange{Wonderful}{10}\hlorange{.”}{12} Sara lowered her voice\hlorange{.}{10}} \\
        } \\ \addlinespace[0.5em]
  & 7089  & Transitional phrases indicating a change in topic or focus.  
        & \makecell[l]{
        \texttt{1.~We\hlorange{~now}{14}\hlorange{~turn}{57}\hlorange{~to}{36} the\hlorange{~facts}{7}\hlorange{~of}{18} this case. The remand ...} \\
        \texttt{2.~We\hlorange{~now}{13}\hlorange{~turn}{58}\hlorange{~back}{39}\hlorange{~to}{33} Pou\hlorange{,}{12}\hlorange{~which}{14}\hlorange{,}{11} like the instant matter...} \\
        \texttt{3.~We\hlorange{~next}{30}\hlorange{~turn}{40} our\hlorange{~attention}{54}\hlorange{~to}{20} the\hlorange{~case}{24}\hlorange{~of}{8} an ample Calabi-Yau} \\
        \texttt{~~~hypersurface  in a complete simplicial toric variety...} \\
        } \\ \addlinespace[0.5em] 
  & 12993 & Elements of societal roles within historical contexts.  
        & \makecell[l]{
        \texttt{1.~<bos> 1812 \hlorange{~Count}{7}\hlorange{~Lie}{17}\hlorange{ven}{21}\hlorange{~was}{20}\hlorange{~appointed}{9}\hlorange{~Ambassador}{15}\hlorange{~in}{10}\hlorange{~London}{11}...} \\
        \texttt{2.~Proper\hlorange{~women}{40}\hlorange{~had}{22} no choice;\hlorange{~they}{26} had to\hlorange{~prevent}{8}\hlorange{~her}{11} acceptance} \\
        \texttt{~~~into \hlorange{~society}{54} as part of their defense...} \\
        \texttt{3.~...from\hlorange{~the}{14} White House that was\hlorange{~a}{15} well\hlorange{-}{19}known\hlorange{~social}{40}\hlorange{~hub}{12}} \\
        \texttt{~~~popular\hlorange{~with}{14} politicians...} \\
        }\\
\bottomrule
\end{tabularx}
\caption{\textbf{Detailed interpretation of exemplar features.} For each SAE feature, we show three illustrative activation snippets. It is important to note two limitations of this interpretability analysis. \textbf{First}, features in an SAE do not map directly to simple natural language concepts; the interpretation provided is an approximation of a feature's function and does not capture its full complexity. \textbf{Second}, this table presents an analysis of single features in isolation. In practice, LLM generation is driven by the combination of many features, and according to the current state of research, it is not yet possible to directly interpret the semantic space formed by such combinations. Future advances in interpretability may address this challenge.}
\label{tb:feature_analysis_supplement}
\end{table*}

\subsection{Logit Lens Analysis}  
\begin{figure*}[t]
  \centering
  \begin{subfigure}[t]{0.48\textwidth}
    \includegraphics[width=\textwidth]{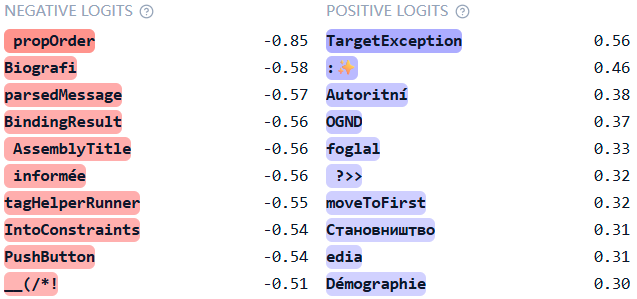}
    \caption{Logits of feature \#755}
    \label{fig:subA}
  \end{subfigure}
  \hfill
  \begin{subfigure}[t]{0.48\textwidth}
    \includegraphics[width=\textwidth]{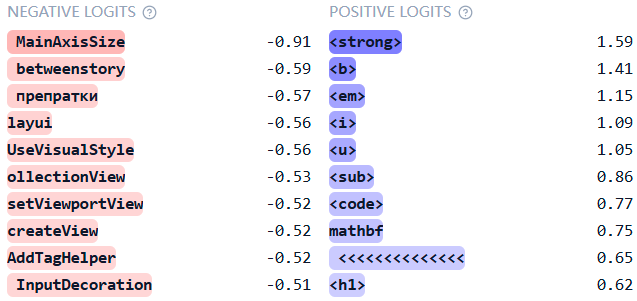}
    \caption{Logits of feature \#10184}
    \label{fig:subB}
  \end{subfigure}

  \vspace{1ex}
  \begin{subfigure}[t]{0.48\textwidth}
    \includegraphics[width=\textwidth]{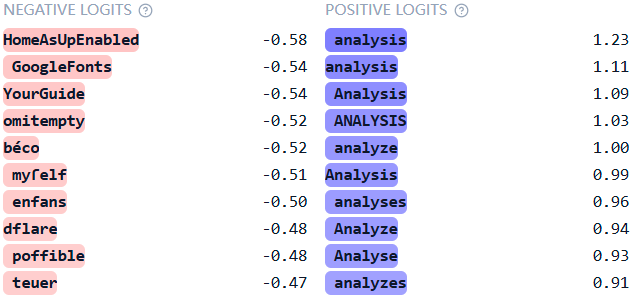}
    \caption{Logits of feature \#11848}
    \label{fig:subC}
  \end{subfigure}
  \hfill
  \begin{subfigure}[t]{0.48\textwidth}
    \includegraphics[width=\textwidth]{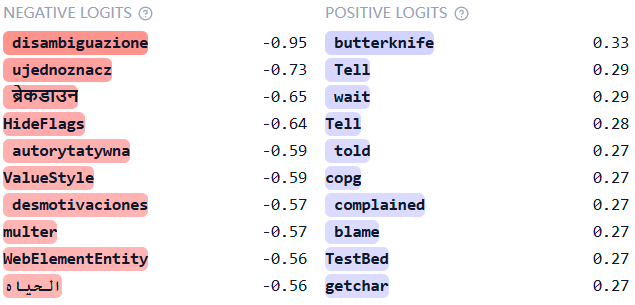}
    \caption{Logits of feature \#15445}
    \label{fig:subD}
  \end{subfigure}

  \vspace{1ex}
  \begin{subfigure}[t]{0.48\textwidth}
    \includegraphics[width=\textwidth]{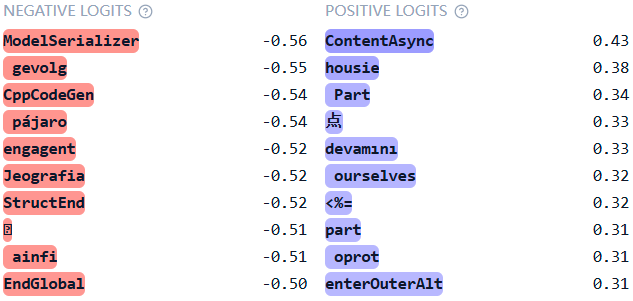}
    \caption{Logits of feature \#7089}
    \label{fig:subE}
  \end{subfigure}
  \hfill
  \begin{subfigure}[t]{0.48\textwidth}
    \includegraphics[width=\textwidth]{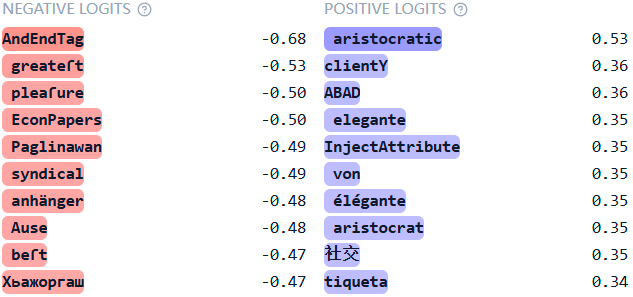}
    \caption{Logits of feature \#12993}
    \label{fig:subF}
  \end{subfigure}

  \caption{Activation profiles for six representative SAE features. In each subfigure, the top negative logits (red) identify tokens most inhibited by the feature, while the top positive logits (blue) highlight tokens most strongly activated. Panels (a)–(f) correspond to features \#755, \#10184, \#11848, \#15445, \#7089, and \#12993, respectively.}

  \label{fig:feature_dist}
\end{figure*}

To understand each feature’s influence on token selection, we present Neuronpedia’s “Logit Lens” analysis~\cite{neuronpedia}, which projects a feature direction \(v_i\) through the model’s unembedding matrix \(W_{\text{unembed}}\). Concretely, for every token \(t\) in the vocabulary, it is computed  
\[
\text{logit}_i(t) \;=\; \bigl(W_{\text{unembed}}\,v_i\bigr)_t,
\]
so that \(\text{logit}_i(t)\) measures the change in the model’s pre-softmax score for \(t\) when feature \(i\) is activated. Over a large evaluation corpus, Neuronpedia aggregate these scores to form a distribution:  
\begin{itemize}
  \item \textbf{Top positive logits (blue):} Tokens with the highest \(\text{logit}_i(t)\), indicating those most strongly promoted by feature \(i\).  
  \item \textbf{Top negative logits (red):} Tokens with the lowest \(\text{logit}_i(t)\), indicating those most suppressed when feature \(i\) is active.  
\end{itemize}
Figure~\ref{fig:feature_dist} visualizes these two sets of tokens for each selected feature, revealing the semantic biases encoded by \(v_i\): positive logits highlight the feature’s preferred concepts, while negative logits expose the concepts it inhibits.

\section{Discussion: Does the Copyrighted Subspace Behave as Expected?}

A critical question is whether our identified subspace $\hat{\mathcal{S}}$ genuinely captures copyright-specific semantics or merely overfits to distributional differences between the copyrighted corpus (news or book) and the general corpus (57 subjects in MMLU). Our experimental results provide two key pieces of evidence against this overfitting hypothesis.

\subsection{Primary Evidence: Our method preserves utility on unseen, in-domain data.} If the subspace were simply a filter for topic (e.g., "news") or text format, its application would harm the model's performance on new tasks from the same domain. However, we observe the opposite. 
\begin{itemize}
    \item As shown in Tables 1 and 2 in main text, while \textsc{SCoPe} significantly reduces infringement, F1 and ROUGE-L scores on new, in-domain NewsQA and BookSum tasks drop by less than 2 percentage points.
    \item This stability is crucial: it demonstrates that SCOPE is not crudely suppressing a topic but is precisely targeting the act of content regurgitation, leaving task-solving capabilities intact.
\end{itemize}
\subsection{Supporting Evidence: The subspace is causally effective, not just a statistical artifact.} A subspace that merely captured a topic or temporal bias would be unlikely to have a predictable, functional role. 
\begin{itemize}
    \item Our reverse intervention experiment (Section 4.3 and Figure 3 in main text) shows a direct causal link: actively amplifying the features in  $\hat{\mathcal{S}}$ consistently increases infringement risk (the win rate drops from 7.8\% to 4.1\%).
    \item This predictable, bidirectional control confirms that the subspace is directly responsible for generating the protected content, rather than being a coincidental byproduct of the data distribution.
\end{itemize}

In summary, because the subspace preserves in-domain utility and is demonstrably causal, we are confident that it controls specific infringement-related semantics rather than simply overfitting to distributional features.

\end{document}